\documentclass[mnsc,rev]{workingpaperJJB} % current default for manuscript submission

\OneAndAHalfSpacedXII
\usepackage{endnotes}
\let\footnote=\endnote

\usepackage{float}
\usepackage{algorithm}
\usepackage{algpseudocode}
\usepackage{multirow}
\usepackage{makecell}
\usepackage{hyperref} % See note above, other options in documentclass may be needed
\usepackage{bbm}
\usepackage{cancel}
\usepackage{amsmath}
\usepackage{graphicx}
\usepackage[font=scriptsize]{caption}
\usepackage{subcaption}
\usepackage{longtable}
\usepackage{xcolor}

\usepackage{natbib}
 \bibpunct[, ]{(}{)}{,}{a}{}{,}%
 \def\bibfont{\small}%
 \def\BIBand{and}%

\TheoremsNumberedThrough     % Preferred (Theorem 1, Lemma 1, Theorem 2)
\ECRepeatTheorems

\EquationsNumberedThrough    % Default: (1), (2), ...
\begin{document}
%%%%%%%%%%%%%%%%

%\setcounter{page}{1}
%\DOI{10.1287/xxxx.0000.0000}%

\RUNAUTHOR{xxxx}

\RUNTITLE{Personalized Financial Incentives for Weight Loss}

% \TITLE{Bundling Information Goods of Decreasing Value}
\TITLE{An Adaptive Optimization Approach to Personalized Financial Incentives in Mobile Behavioral Weight Loss Interventions}

\ARTICLEAUTHORS{%
\AUTHOR{Qiaomei Li, Yonatan Mintz}
\AFF{Department of Industrial and Systems Engineering,UW Madison, Madison, WI, \EMAIL{\{qli449,ymintz\}@wisc.edu}} %, \URL{}}
% \AUTHOR{Yonatan Mintz}
% \AFF{Department of Industrial and Systems Engineering,UW Madison, Madison, WI, \EMAIL{ymintz@wisc.edu}}
\AUTHOR{Kara L. Gavin, Corrine I. Voils}
\AFF{Department of Surgery,UW Madison, Madison, WI, \EMAIL{\{gavin,voils\}@surgery.wisc.edu}}
% \AUTHOR{Kara Gavin}
% \AFF{Department of Surgery,UW Madison, Madison, WI, \EMAIL{gavin@surgery.wisc.edu}}
} % end of the block

\ABSTRACT{%
Obesity is a critical healthcare issue affecting the United States. The least risky treatments available for obesity are behavioral interventions meant to promote diet and exercise. Often these interventions contain a mobile component that allows interventionists to collect participants level data and provide participants with incentives and goals to promote long term behavioral change. Recently, there has been interest in using direct financial incentives to promote behavior change.  However, adherence is challenging in these interventions, as each participant will react differently to different incentive structure and amounts, leading researchers to consider personalized interventions. The key challenge for personalization, is that the clinicians do not know a priori how best to administer incentives to participants, and given finite intervention budgets how to disburse costly resources efficiently. In this paper, we consider this challenge of designing personalized weight loss interventions that use direct financial incentives to motivate weight loss while remaining within a budget. We create  a machine learning approach that is able to predict how individuals may react to different incentive schedules within the context of a behavioral intervention. We use this predictive model in an adaptive framework that over the course of the intervention computes what incentives to disburse to participants and remain within the study budget. We provide both theoretical guarantees for our modeling and optimization approaches as well as demonstrate their performance in a simulated weight loss study. Our results highlight the cost efficiency and effectiveness of our personalized intervention design for weight loss.

% Enter your abstract
}%

%\KEYWORDS{deterministic inventory theory; infinite linear programming duality;
%  existence of optimal policies; semi-Markov decision process; cyclic schedule}

 %\HISTORY{This paper was
%first submitted on April 12, 1922 and has been with the authors for
%83 years for 65 revisions.}

\maketitle
%%%%%%%%%%%%%%%%%%%%%%%%%%%%%%%%%%%%%%%%%%%%%%%%%%%%%%%%%%%%%%%%%%%%%%

% Samples of sectioning (and labeling) in OPRE
% NOTE: (1) \section and \subsection do NOT end with a period
%       (2) \subsubsection and lower need end punctuation
%       (3) capitalization is as shown (title style).
%
%\section{Introduction.}\label{intro} %%1.
%\subsection{Duality and the Classical EOQ Problem.}\label{class-EOQ} %% 1.1.
%\subsection{Outline.}\label{outline1} %% 1.2.
%\subsubsection{Cyclic Schedules for the General Deterministic SMDP.}
%  \label{cyclic-schedules} %% 1.2.1
%\section{Problem Description.}\label{problemdescription} %% 2.

% Text of your paper here

\section{Introduction}

The obesity epidemic is one of the most critical health issues facing the United States. According to the adult obesity data in 2017-2020 from the Center for Diseases and Prevention (CDC), the prevalence of obesity is 41.9\%  \citep{stierman2021national}. Obesity increases the risk of metabolic diseases such as type 2 diabetes and heart disease \citep{golay2005link} and has led to related medical costs of \$173 billion in the United States in 2019 \citep{ward2021association}. If an individual with obesity is able to achieve a moderate reduction in weight (by  5\%), they can mitigate many of these adverse effects \citep{wing1987long,krentz2016evolution}. Currently, the lowest risk treatments that have been found to be effective for treating obesity involve clinically monitored behavioral interventions \citep{grilo2011cognitive, jakicic2016effect}.  Given advances in technology, recent generations of these interventions include a mobile application as a component in which individuals are asked to record their daily weight, exercise, and or daily calories consumed \citep{fukuoka2018applying}. These applications can be used to provide participants with feedback and rewards to encourage behavioral change and weight loss. One key challenge in these interventions is that participant adherence decreases over time \citep{acharya2009adherence, lemstra2016weight}. 

Several studies have shown that financial incentives for weight loss could improve adherence and lead to clinically significant weight loss of at least 5\% of baseline weight \citep{john2011financial, volpp2008financial}. The primary objective of the intervention in these studies is to maximize the number of participants who achieve clinically significant weight loss at the end of the study \citep{wing1987long}. To achieve this goal, the interventionist can dispense financial incentives to each participant to encourage weight loss and calorie recording. Previous studies have compared different reinforcement schedules, amounts, and targets in an attempt to determine the optimal structure on average of a financial incentives intervention. In these previous studies, incentives have followed a predetermined treatment schedule that does not adapt to participant data collected over the course of the intervention \citep{tsai2020mobilizing}. In other words, all participants can receive the same amount of money for achieving the same criteria (e.g., weight loss, calorie recording). One key challenge for the interventionist in this setting is that each individual participant will have different levels of motivation stemming from financial incentives and internal desire to lose weight, leading to heterogeneity of response to financial incentives. Moreover, these individual motivations are unknown to the interventionist \textit{a priori}, and must be inferred from participant data. A second challenge is that, to ensure the total intervention costs are manageable, the interventionist can only disburse incentives from some maximum total intervention budget. Typically, this budget is distributed evenly across all participants, such that each participant can earn a maximum amount. Accordingly, when scaling
up the intervention to more participants, the cost increases linearly. A key operational challenge that must be addressed is how to design a modeling and optimization framework that can allow the interventionists to disburse costly incentives to match individual motivations and encourage the largest number of participants to lose weight.

In this paper, we propose a novel optimization framework that addresses these key challenges. While existing work in the operations literature has modeled how individuals respond to indirect motivational goals such as exercise goals and messaging \citep{aswani2019behavioral,mintz2020nonstationary}, in this paper we focus on modeling how participants respond to direct monetary incentives for weight loss. Our proposed framework involves first providing a behavioral model for participants in the context of weekly financial incentives. This model is meant to capture the dynamics of how participant motivational states (i.e., intrinsic and extrinsic motivations) change over time and how they impact their choices with regards to calorie consumption and recording their physical state (i.e., their weight). We then propose a surrogate likelihood approach for estimating these unknown participant state parameters and provide an approach to use these estimates to predict future participant response to potential interventions. The last step of our framework involves using this predictive model to optimize the incentives awarded to each participant. We provide an adaptive algorithm that can calculate an asymptotically optimal incentive policy while staying with in the financial resource constraints of the intervention. Our adaptive approach can be calculated weekly over the course of a weight loss intervention to improve its estimation using data obtained from participants currently participating in the trial, and compute new incentives that adapt to changing study conditions or participant response.% We validate our methodology using a computational study based on data from a previous randomized control trial (RCT) that provided incentives of a similar structure.  

\subsection{Clinical setting: the Log2Lose study}

We developed our modeling and optimization methods using the data and structure from a study that investigated the impact of different financial incentive structures on weight loss called Log2Lose \citep{voils2018study}. The 24-week Log2Lose pilot trial
was designed to evaluate the feasibility of delivering incentives in near
real-time using data collected from cellular-connected scales and a mobile food and activity tracking. 
%In previous studies, receipt of incentives has been delayed due to requiring people to attend in-person visits to provide data. 
The goal of Log2Lose is to compare the efficacy of incentives for two different outcomes, either individually or combined: weekly calorie recording or weekly weight loss. Accordingly, participants were randomized to one of four arms: incentives for both calorie recording and weight loss (Arm A); incentives for calorie recording (Arm B); incentives for weight loss (Arm C); or no incentives (Arm D). The incentive schedule was based on psychological learning theory and involved the following principles: 1) It was fixed at \$10 for the first four weeks to encourage learning of new behaviors, and 2) It varied between \$0 and \$30 per week thereafter. Thus, even if
participants had the desired outcome, they did not receive a reward some weeks. The
predetermined incentive scheduled applied to all participants. It was not known by participants \emph{a priori} and thus appeared random.

All participants were invited to a biweekly group counseling session led by a registered dietitian that involved dietary education and behavioral skills such as regular self-weighing and calorie recording. Participants were encouraged to record a minimum amount of calories (1,000 KCal for females and 1,200 KCal for males) in a mobile application for at least five days a week, with at least one of the days being a weekend day. Daily calories recorded were transmitted to the research team through an open application programming interface. If participants in Arm B met this goal, they were awarded a monetary payment between \$0-\$30. Additionally, all participants were given a cellular scale that transmitted their weight measures to the investigators whenever they weighed themselves. They were encouraged to weigh themselves at least two times each week. The difference between the first and last weight of the week was taken. Participants in Arm C received a payment between \$0-\$30 each week that the last weight was lower than the first weight. Arm A combined both weight loss and caloric recording incentives but reduced the reward range to \$0-\$15 for each to ensure the maximum amount that a participant could earn was \$30 a week. The control arm did not receive any financial incentives. Recorded calories and weight were compiled and analyzed in a software application, and notice of incentives was provided using text messaging. 
%Weekly incentives for Arms A-C were disbursed on a debit card. 
For analysis in this paper, we used the cellular weight data, app data on calorie recording, the record of awarded incentives, as well as participant demographics for the purposes of validating our models and conducting our numerical studies. For the full demographic data and trial protocols please see \citep{voils2021randomized,voils2018study}. We note that, while our model is based on the structure of this particular intervention, we believe the approaches and techniques we develop will be widely applicable to other behavioral interventions developed in the future that may have direct financial incentive components.

\subsection{Related literature}
While our modeling is based in the clinical setting of behavioral interventions, through our modeling and optimization analysis we  contribute to three streams of literature within the operations field. These include sequential decision making methods\ref{sec: or methods}, healthcare operations research  \ref{sec: seq decision-making}, and predictive modeling for clinical weight loss \ref{sec: wl prediction}.

\subsubsection{Sequential decision making methods} \label{sec: or methods}
Our setting of computing weekly individual level financial incentives for participants fits generally into the stream of sequential decision making methods with partial information. In particular we can think of our setting as that of a decision maker (the interventionist) taking sequential control actions (weekly incentives) with respect to a system state for which they have imperfect information (participant motivations and weight). Two of the main approaches for addressing the problem of making sequential decisions with imperfect information include  partially observable Markov decision process (POMDP) \citep{yu2008near, ayer2012or} and reinforcement learning (RL) \citep{sutton2018reinforcement}. The key difference between these families of approaches is that, in the POMDP setting, the decision maker is assumed to have partial information of the system state while having information of the system dynamics; in contrast, in RL the decision maker is assumed to have full information of the system state while having partial information of the system dynamics.

The classical solution technique used for POMDP models involves reformulating the POMDP as what is known as a belief Markov Decision Process (MDP), by considering what is known as a belief state, a state that encodes the decision maker's belief they are in any of the POMDP states \citep{bertsekas2012dynamic}. In general, the belief state can be thought of as a distribution over the state space of the POMDP that reflects likelihood of a particular state being the true state of the system at any given point in time. However, solving the belief MDP in practice is quite challenging since even if the state space is finite, the belief state could be uncountably infinite. Therefore, in the POMDP literature different techniques such as approximate dynamic programming (ADP) \citep{yu2012discretized, dai2019inpatient} and policy gradient \citep{zhang2021convergence} have been used for approximating the optimal solutions. Our setting can be thought of as a particular instantiation of a POMDP, with specific model structure. Using our model structure, we develop an approximate solution method that is asymptotically optimal under a set of mild conditions.

%In contrast, instead of targeting a decision process with discrete state and discrete action space we focus on a continuous state and continuous control problem. In addition, we pursue a model-based approach to 
%
%have been widely used for solving sequential decision making problems, and the main difference between them is a POMDP is often implemented to solve problems with unknown states and reinforcement learning (RL) is implemented to learn unknown dynamics.
%
%A classical way to solve a POMDP is to reformulate the problem as a Markov decision process (MDP) with belief state \citep{bertsekas2012dynamic}, which is a probability distribution indicating the probability of the system in each possible state. However, even if the state space is finite, there are infinitely countable choices for the belief state, which makes the computation often intractable in practice. Therefore, different techniques such as approximate dynamic programming (ADP) \citep{yu2012discretized, dai2019inpatient} and policy gradient \citep{zhang2021convergence} have been used for approximating the optimal solutions. In contrast, instead of targeting a decision process with discrete state and discrete action space we focus on a continuous state and continuous control problem. In addition, we pursue a model-based approach to 

Methods in the RL literature can be categorized into two broad families namely model-based RL \citep{zhou2018personalizing, osband2014model}, which use specific functional form (or parametric estimates) of the transition dynamics and value function, and model-free methods \citep{strehl2006pac, akrour2016model}, which use stochastic approximations of the problem value functions and transition probabilities without explicit functional forms. In this paper, our proposed approach can be thought of as a form of model-based RL as we explicitly model system dynamics (e.g., dynamics of participant weight and motivations). Our modeling framework is related to existing model-based methods developed for behavioral weight loss interventions \citep{mintz2017behavioral, zhou2018personalizing}.

\subsubsection{Healthcare operations research} \label{sec: seq decision-making}
Our setting is related to the large stream of existing work on applications of operations research to healthcare applications. In particular there has been a vast amount of work examining applications of sequential decision making in managing the operations of providing care \citep{ekici2014modeling,erdogan2013dynamic,childers2009prioritizing}, providing personalized treatment \citep{ayer2019prioritizing,bastani2020online,schell2016data,mintz2020nonstationary,he2023model}, and intervention management \citep{deo2013improving,lee2019optimal}. Our problem setting and methods contribute to these streams of literature, in particular to the work focused on personalized treatment and intervention management. Much like these settings, we consider a resource constrained problem, where decision makers must make costly decisions under uncertainty. One of the contributions of our work is in developing a framework that extends the existing work in these settings to behavioral interventions where a decision maker must disburse financial incentives to participants with imperfect information. In contrast to existing work that considers resource constraints on manpower or facilities, our work examines a constraint on the direct budget of the intervention and how it can be best disbursed amongst participants to motivate them to achieve weight loss.

\subsubsection{Predictive models for weight loss } \label{sec: wl prediction}
 Our work is also related to a stream of literature that focuses on predicting an individual's weight loss success in the context of a clinically supervised intervention. Existing predictive models for this setting include differential equations \citep{thomas2011simple}, Markov models \citep{bromberger2014weight}, data mining methods \citep{batterham2017using}, and machine learning methods \citep{lee2020multi}. In general, these methods were developed to perform a binary prediction task (i.e. whether or not a participant achieves clinically significant weight loss), making them challenging to use for optimization. In contrast, the behavioral framework we develop in this paper is capable of providing predictions for the full weight trajectory of study participants given a particular sequence of financial incentives. Our framework is also able to compute the likelihood of such a trajectory occurring and can thus also be used for binary prediction in addition to this regression task in a similar manner to \cite{aswani2019behavioral}. However, our work  differs from the predictive approach in \cite{aswani2019behavioral} in two key ways. First, we focus on a weight loss intervention with financial incentives instead of motivational goals, which are evaluated by participants in a slightly different manner and thus alter the model structure. In particular, the nature of the weekly financial incentives means participants value their actions over the course of several days (during the incentive evaluation period), unlike daily step goals that only impact participant behavior during a single day of the study. Second, our model incorporates both continuous and discrete measurements, making it challenging to use maximum likelihood estimation directly. We propose to solve this challenge using surrogate likelihood estimation, a more challenging method to analyze.

\subsection{Contributions}
In this paper, we develop a framework to design personalized financial incentives that encourage weight loss, while ensuring that intervention costs remain within a fixed budget. Through the development of this framework we make three key contributions:
\begin{enumerate}
    \item We extend participant behavioral models in weight loss interventions to capture the effect of financial incentives on participant behavioral change. Our novel modeling additions include both medium and long term impacts of financial incentives, and capture how repeated use of financial incentives may not lead to meaningful long-run behavioral change. In particular, we are able to capture both short-term (in-week) participant decisions as well as long-term (between-weeks) participant behavioral change. Our model incorporates insights from self-determination theory (SDT), namely that it includes parameters both intrinsic and extrinsic motivation for weight loss. According to SDT, motivation ranges on a
continuum from completely nonself-determined (lacking motivation) to self-determined
(intrinsically motivated); in between are several levels of extrinsic motivation in which one’s behavior can be completely or partially driven by external sources such as rewards and punishment \citep{deci2013intrinsic}.
    \item We develop a novel inverse optimization approach for estimating unknown participant parameters and states that is statistically consistent. In contrast to existing literature which looked at inverse optimization for purely myopic participants \citep{mintz2017behavioral}, our approach assumes participant's plan for the medium-term using dynamic programming, and uses the structure of the resulting optimal policy to construct a set of constraints for inverse optimization. We then use these constraints in a  surrogate likelihood estimation model, which can be solved using commercial mixed integer programming solver. We further show the resulting estimates are statistically consistent, which, to our knowledge, is one of the first consistency guarantees shown for surrogate likelihood models trained with non-convex optimization. Furthermore we show how these estimated parameters can be used in an adaptive optimization framework to allocate incentives for weight loss given a budget that we call the Design of Incentives Algorithm (DIA). Through theoretical analysis, we show that the incentive policy output by DIA is asymptotically optimal.
    %\item We extend the behavioral analytics algorithm for constrained financial incentives that can take budget, and we show the resulting estimated financial incentives are asymptotically optimal under the deterministic incentive policy.
    \item We conduct a comprehensive set of numerical validation studies using real-world-data from the Log2Lose trial, which deployed financial incentives to help participants achieve clinically significant weight loss. Our experiments demonstrate that our proposed behavioral model is descriptive of participant behavior, and moreover is capable of better predictive performance than existing state-of-the-art machine learning approaches. Furthermore, through a simulation study we are able to show that our dynamic optimization framework is able to achieve improved clinical outcomes for less budget when compared to existing one-size-fits-all approaches, indicating that using our methods such interventions could be scaled to larger participant populations. 
\end{enumerate}

\section{Participant behavioral model}\label{sec:pat_prob}
% In this section we describe a participant's decision-making process in a weight loss intervention using a dynamic programming model.
 %
 Here, we present our model for participant behavior during a weight loss intervention. We use a utility maximization framework where participants are assumed to make weight loss-related decisions (namely how many calories to consume each day and whether or not to record their calories) based on individual utility functions that depend on their perceived health benefits, their responsiveness to financial incentives, and preferred level of caloric consumption. Our model consists of three key classes of variables we call physical system states, which are state variables that capture the physical aspects of weight loss (namely the participant's weight), motivational states that capture a participant's cognitive state and how much importance they place on different actions and health outcomes (i.e., intrinsic and extrinsic motivation for weight loss gained from financial incentives), and decision variables that represent a participant's actions that affect weight loss (i.e., daily caloric intake). A key feature of our model is that all physical and motivational states are modeled as individual specific, and thus will be different for each participant. Because of this, we focus our modeling discussion on modeling the behavior of a single participant.

 To capture how participant behavior changes over time as a consequence of the intervention, we also define a set of dynamics that describe how the motivational and physical states change over the course of the program as a consequence of the intervention treatment and individual participant decisions. Since financial incentives are administered to the participant based on their weight loss and calorie recording at the end of a study week our framework models the participant's decision-making process in two components: 1) A component that models the participant's daily actions over the course of a single week of a trial given their expectation for financial reward, we call this component the in-week decision model. 2) A component that models long term behavioral change by tracking how participant motivational states change as a consequence of the previous week's actions and financial incentives; we call this component the between-week decision model. Both of these time frames are fully integrated into a single participant model, which, as previously noted, is individual-specific and captures the unique way each participant will interact with the intervention. A key assumption to these models is that participants make decisions in a myopic utility-maximizing manner, that is, they only make decisions on calorie consumption during the course of a study week that will impact their financial incentive earned for that week and will not consider future incentives or long term health benefits. This behavior has been observed in the social science literature, and can be framed as participants making rational decisions with respect to high future discounting of health and monetary gain \citep{cawley2004economic}. Prior work has shown that models that incoroporate this assumption still provide strong predictive and descriptive performance  \citep{aswani2019behavioral,mintz2017behavioral,adams2023planning}. We note that while existing myopic models consider participants that only consider single daily decisions, due to the structure of the financial incentives in our setting, the myopic assumption implies participants consider their decisions at the start of a week.

\subsection{Participant in-week decision model}

The first step of our framework is to describe the participant's daily decision making process during a single study week. Let $t$ be the week index and $d \in \{0,...6\}$ be the day index where each week starts on Monday ($d=0$) and ends on Sunday ($d=6$). Let the physical system states $w_{t,d},f_{t,d} \in \mathcal{W} \times \mathcal{F}$ be the participant's weight and caloric consumption on day $d$ of week $t$, where $\mathcal{W}, \mathcal{F} \subset \mathbb{R}_+$ are closed intervals. Let the motivational states of the participant be given by $a_{1,t}, a_{2,t},f_{b,t} \in \mathcal{A}^2 \times \mathcal{F}$  that represent the participant's internal motivation, that is a participant's personal motivation for weight loss, external motivation for weight loss, or how influential financial incentives are on the participant's motivation to lose weight, and the participant's preferred caloric consumption level on week $t$ respectively. Here $\mathcal{A} \subset \mathbb{R}_+$ is assumed to be a closed interval. The participant's decisions in this model are denoted by $c_{t,d} \in \mathcal{F}$ that represent the participant's planned caloric intake on day $d$ of week $t$. Note that unlike existing models that consider caloric consumption directly as a participant decision \citep{aswani2019behavioral,mintz2017behavioral}, a key feature of our model will be that, while participants are capable of planning to a particular value of caloric consumption, this may not equal the amount of calories they truly consume. This is a challenge in many calorie-recording based behavioral interventions since, even when trying to the best of their abilities, participants often cannot accurately estimate the amount of calories they consume with each meal \citep{mckenzie2021investigating}. Furthermore, social desirability concerns may encourage under-reporting of caloric consumption. The final component of the in-week decision model is a motivational state that captures the participant's expectation for financial incentives at the end of the week. We denote the amount of financial incentive allocated by the interventionists for weight loss at the end of week $t$ by $r^w_t \in \mathcal{R}$, where $\mathcal{R} \subset \mathbb{R}^+$ is a closed interval. However, since this amount is generated at the end of the week based on the participant's performance and the intervention is structured so that financial incentives seem randomly generated to the participant conditioned on meeting the goal, individuals cannot use the true value of the incentive for their decisions during week $t$. Instead participants form a belief on the financial reward they will potentially receive at the end of the week should they meet their weight loss goal based on their previous rewards received and knowledge of the intervention policies. We let $\hat{r}_t^w \in \mathcal{R}$ be a random variable that represents the participant's estimate of their potential financial reward for weight loss in week $t$ that influences their decisions based on these beliefs.

%Next, let $\{a_{1,t}, a_{2,t}\}$ be the set of motivational states that represent the participant's internal motivation, that is a participant's internal cognition and personal motivation for losing weight,  and external motivation for weight loss, that is how influential financial incentives are on the participant's motivation to lose weight, respectively.
%
%We denote the participant's main decisions with the variable $c_{t,d}$ that represent the participant's planned caloric intake on day $d$ of week $t$.  Let $f_{t,d}$ be the preferred caloric intake on day $d$ of week $t$ and $f_{b,d,t}$ be the baseline value of preferred caloric intake.
%
%Let $r^w_w$ denote the weight loss incentive of week $w$ and $r^c_w$ denote the calorie recording incentive of week $w$.

Using these variables, we model the participant's in-week decision process for week $t$ of the intervention as the following utility maximization problem, 
\begin{subequations}
 \label{eq:in_week_model}   
\begin{align}
\max_{\{c_{t,d}\}_{d=0}^6} \ \mathbb{E}&\big[ -a_{1,t}\sum_{d=1}^6 w_{t,d}
+a_{2,t}\hat{r}^w_{t} \mathbbm{1}\{w_{t,0}- w_{t,6}>0\}
- \sum_{d=0}^6(f_{t,d} - f_{b,t})^2\big] \label{eq: in week 0}\\
 \text{subject to:} \ 
& w_{t,d+1}=bw_{t,d}+cf_{t,d+1}+k, \quad d \in \{0,\cdots,5\}, \label{eq: in week 1}\\
%& \tilde{w}_{w,d} = w_{w,d}  + \zeta_t, \quad d \in [0,1,\cdots,6]\\
&f_{t,d}=c_{t,d}+\xi_d, \quad  d \in \{1,\cdots,6\}\label{eq: in week 2},\\
%&f_{b,t,d} = \gamma(f_{b,t,d-1} - f_{b,t}) + f_{b,t}. &  d \in \{1,\cdots,6\} \label{eq: in week 3}
& w_{t,d},f_{t,d}, c_{t,d} \in \mathcal{W}\times \mathcal{F}^2, \quad d \in \{0,...,6\}.
\end{align}
\end{subequations}
The interpretation of this model is that the participant's planned caloric intake at each day $d$ of week $t$ is given by the argmax of the above optimization problem where the objective given by \eqref{eq: in week 0} represents the participant's utility function and \eqref{eq: in week 1}-\eqref{eq: in week 2} represent the dynamics of the participant's weight and caloric intake preferences. Note that \eqref{eq: in week 0} contains three main components that impact the participant's decisions. The first term $-a_{1,t}\sum_{d=1}^6 w_{t,d}$ indicates that the participant wants to reduce their future weight for each day of the week, and this is weighted by their motivation for weight loss $a_{1,t}$. The next term $a_{2,t}\hat{r}^w_t \mathbbm{1}\{w_{t,0} - w_{t,6} > 0\}$ indicates participants would like to reduce their weight over the course of the week so that they can be eligible for the financial reward, and this is weighted by $a_{2,t}$ that expresses how motivated they are by financial rewards. The final term $-\sum_{d=0}^6 (f_{t,d} - f_{b,t})^2$ indicates that participants want to choose foods with calories each day
that are close to a certain caloric preference level $f_{b,t}$. This last component signifies that, without intervention, there exists some theoretical preferred amount participants would desire to eat that is not so little that they would feel hungry or so much that it would be physically impractical. Constraint \eqref{eq: in week 1} represents the dynamics of weight loss using the Mifflin St. Jeor equation \citep{mifflin1990new} where $b,c$ are known constants and $k$ is a constant computed from the participant's age, gender, and height. Constraint \eqref{eq: in week 2} models that despite planning to consume $c_{t,d}$ participants may over or under eat since they cannot get an accurate estimate of their calories. This uncertainty is captured by i.i.d. disturbance variables $\xi_{t,d}$, that we assume are bounded such that $f_{t,d} \in \mathcal{F}$ with probability of one and $\mathbb{E}\xi_{t,d} = 0$. Specifically we assume $\xi_{t,d} \sim U(-A,A)$, in other words that the deviation from the calorie plan is uniformly distributed within $A$ calories. While other distributions could be used to model this uncertainty, we chose the uniform distribution for computational reasons to enable us to estimate the unknown model parameters using commercial mixed integer programming (MIP) solvers, this reformulation is described in detail in Section \ref{sec: mle}. %The final constraint \eqref{eq: in week 3} captures that participant's caloric consumption preferences will tend towards a new weekly baseline as the week progresses. Here $\gamma \in (0,1)$ is a decay rate that captures how fast the preferences converge to the new baseline. This constraint can be interpreted as showing that long term change in caloric consumption behavior will manifest in daily decisions fully only after the participant has practiced the new behavior for an extended amount of time.

\subsection{Participant between-week model}
\label{sec:pat_bet_wek_mod}
Next, we describe the model for how participant behavior evolves from week to week. While over the course of a single week participants do not respond directly to the financial incentives (since they are awarded at the end of the week) this model captures how weekly incentives change participant motivation over the course of the intervention. Therefore, unlike the in-week model, all components of this model describe the evolution of motivational states and not physical states or decisions.

Using the previous notation let $a_{1,t},a_{2,t}$ describe the participant's internal motivation for weight loss and external motivation for weight loss from financial incentives on week $t$, and let $f_{b,t}$ represent the participant's preferred caloric intake on week $t$. Let $g_t$ be an indicator variable that equals 1 when the participant successfully meets their calorie recording goal on week $t$. We model $g_t$ as a Bernoulli random variable since different exogenous influences (such as participants not having time during the week or getting distracted) can influence whether or not they record calories \citep{raber2021systematic}. Let $p_t \in \mathcal{P}\subset (0,1)$ represent the probability a participant will meet their calorie recording goal on week $t$ (that is $p_t = \mathbb{E}g_t$). 

 We define the following set of dynamics that describes the transitions of motivational states $a_{1,t}, a_{2,t}, p_t, f_{b,t}, \hat{r}_t^w$:%. Recall $a_{1,t}, a_{2,t}$ are the motivation for weight loss and weight loss incentive. Let $p_t$ be the probability of the participant satisfying the calorie recording requirement in week $t$, and let $f_{b,t}$ be the baseline of preferred caloric intake in week $t$.
 \begin{align}
& a_{1,t+1} = \gamma_1(a_{1,t} - a_{1,b}) + a_{1,b} + k_{1} \mathbbm{1}\{(w_{0}- w_{6})>0\} +  r^c_t\mathbbm{1}\{p_t-B\geq 0\}, &  t\in \{0,\cdots,23\}, \label{eq: bet week 1}\\
 & a_{2,t+1} = \gamma_2(a_{2,t} - a_{2,b}) + a_{2,b} + k_2 r^w_t \mathbbm{1}\{(w_{0}- w_{6})>0 \}, &  t\in \{0,\cdots,23\}, \label{eq: bet week 2}\\
&p_{t+1}=\gamma_p(p_t-p_{b})+p_{b}+k_pg_t, &  t\in \{0,\cdots,23\}, \label{eq: bet week 3}\\
&f_{b,t+1} = \gamma_f f_{b,t} + (1-\gamma_f)\frac{1}{7}\sum_{d=0}^6 f_{t,d}, &  t\in \{0,\cdots,23\}, \label{eq: bet week 4} \\
& \hat{r}_{t+1}^w = \begin{cases} \frac{t}{t+1}\hat{r}_t^w + \frac{1}{t+1}r_t^w, \quad \text{ if } w_0 -w_6 < 0, \\ \hat{r}_t^w, \quad \text{ otherwise, }  \end{cases} & t \in \{0,...,23\}. \label{eq:bet_week_reward_belief}
\end{align}

 %Here, $\gamma_1, \gamma_2,\gamma_p, \gamma_f \in (0,1)$ are the rate of the states returning to their corresponding baseline values 
%
The interpretation of these dynamics is that all motivational states have some baseline values that changed as participants interact with the intervention, but that, as time progresses, the impact of the intervention decays exponentially and the states tend to their baseline. Here, $a_{1,b},a_{2,b},p_{t,b}$ represent the baseline value of each motivational state, which can be interpreted as the motivational states of the participant without any interaction with the behavioral intervention, and $\gamma_1, \gamma_2,\gamma_p \in (0,1)$ are the decay rates at which the states return to baseline. %As time progresses from the last distributed financial incentives, the impact of the incentives on the participant's decision making diminishes and eventually the motivational states return to their baseline conditions.%
$k_1, k_2 \in \mathcal{K}$ represent the increase in motivational states when participants meet their weight loss goal and receive financial incentive respectively, where $\mathcal{K} \subset \mathbb{R}^+$ is a closed interval. $a_{1,w+1}$ increases by $k_1$ if the participant satisfies the weight loss requirements in previous week. This models that participants will be more motivated internally to meet the weight loss goal as they succeed initially in losing weight. Likewise $a_{2,t+1}$ increases by $k_1r^w_t$ if the participant satisfies the weight loss requirements in week $t$ and receives financial incentive $r^w_t$. This would indicate that if a constant positive reward is given to the participant their motivation from financial incentives will increase rapidly. But in cases where $r^w_t = 0$ and the participant still manages to lose weight, only  $a_{1,t}$ will increase while $a_{2,t}$ will return to baseline. This interaction in the dynamics ensures that, in order to impact long-term behavioral change and reduce dependence on incentives, effective policies should at some points provide zero reward even if a participant is likely to lose weight. This notion is well known in the behavioral literature and can be thought of as encoding the principle of intermittent reinforcement \citep{ferster1957mixed}. $B \in \mathcal{P}$ can be interpreted as the baseline probability of a participant satisfying the calorie recording requirements, and $a_{1,t+1}$ only increases when $p_t > B$, that is if the participant is motivated enough to record their calories that this would also reflect on their motivation to lose weight. $k_p$ can be thought of as a parameter encoding the intrinsic motivation of the participant from calorie recording since $p_{t+1}$ increases by $k_p$ if the participant satisfies the calorie recording requirements in week $t$. Moreover $r_t^c$ represents the amount of financial incentive awarded for meeting the calorie recording goal, and its inclusion in \eqref{eq: bet week 1} signifies that if participants are rewarded for calorie recording this will increase their motivation for weight loss in the coming week.

There are two exceptions to these dynamics descriptions. The first is \eqref{eq: bet week 4},  which describes the long-term behavioral change of baseline caloric preference. Essentially, this equations states that future caloric consumption preference can be thought of as a geometric average of the previous caloric preference and the average caloric consumption in the previous week. Thus as participants modify their behavior and have lower weekly consumption this will result in a slow but long term change in the baseline caloric consumption preference of the participant. The second is \eqref{eq:bet_week_reward_belief}, which indicates that participants estimate their expected reward as an arithmetic average of past rewards received so long as they've met the weight loss goal. In other words, if they did not meet the goal (and thus expected to receive zero reward) this belief does not update; however, if they do meet the goal but receive zero reward their reward belief decreases. This means that although providing a reward of zero would be beneficial for long run behavioral change, it could lead to a decrease in weight loss motivation in the short term presenting an important trade-off to the decision maker.
 
%we develop a model that describes the agent’s decision-making process using a set of utility functions, time-varying motivational states and system states. Next, we get estimated model parameters using data from the Log2lose study, which is a randomized trial of weight loss intervention, in which participants receive weekly financial incentives for self-monitoring their diet and weight loss. Then we use our estimated model to solve the incentives allocation problem with limited budget. We find our method outperforms some existing machine learning methods in terms of weight loss prediction at the end of the intervention using a short time-span of data. We also run a simulation study which shows our approach helps improve the efficacy of the intervention program with significant cost reduction.

%To simplify the participant decision-making model, set $a_{4,w}=1 $ and the optimal solution of planned caloric intake $\{c^*_{w,d}\}$ for $d=\{0,1, \cdots, 6\}$ only depend on the motivational states $\{a_{1,w}, a_{2,w}, p_w\}$. Define $p_w$ as the probability of the participant satisfies the calorie recording requirements of week $w$, and let $g_w$ be a binary variable indicating if the calorie recording requirements are fulfilled in week $w$. The dynamics of the motivational states are:

\section{Estimation and prediction of unknown parameters}\label{sec: mle}
While the model described in Section \ref{sec:pat_prob}  is able to capture mathematically the decision making process of participants and their interaction with the intervention, in practice most of the parameters in this model are not known \textit{a priori} to the interventionist. In order to provide effective incentives to individuals so that they can lose weight, the interventionists must be able to estimate these individual level participant parameters using data collected through the intervention. This data comes in two main forms, observations of whether or not participants managed to meet their calorie recording goals on week $t$ ($g_t$ in the notation from Section \ref{sec:pat_prob}) and noisy observations of weight at each day of the intervention that we call $\tilde{w}_{t,d}$. This estimation problem poses two main challenges, namely that the data is noisy, and there could be a significant amount of missing data. This second challenge is of particular interest to the Log2Lose case since in this intervention all weight is self generated through participants using a cellular scale. Depending on various factors (such as how busy their day was or if they are traveling) they may not weigh themselves every day throughout the intervention. 

To address these challenges, we use a joint parameter estimation approach that formulates the estimation problem as a mixed integer program (MIP). Specifically, we consider an approach similar to \cite{aswani2019behavioral} by using a joint maximum likelihood estimation (MLE) approach.  We assume that $\tilde{w}_{t,d} = w_{t,d} + \epsilon_{t,d}$, where $\epsilon_{t,d}$ are i.i.d. noise terms such that $\mathbb{E}\epsilon_{t,d} = 0$ and $\mathbb{E}\epsilon_{t,d}^2 = 0$. For our specific formulation  we assume that $\epsilon_{t,d} \sim \text{Laplace}(0,\sigma)$, but note that our analysis could apply to all noise distributions that can be represented using a set of mixed integer linear constraints such as piece-wise linear distributions or the shifted exponential distribution.This is formalized by the following proposition.
\begin{proposition}\label{prop: mle formulation}
 The MLE problem can be formulated as the following constrained optimization problem:
\begin{subequations}
\begin{align}
\max_{\{w_{t,d}, p_t,B , a_{1,t},a_{2,t}, f_{b,t},f_{t,d},c_{t,d}, \hat{r}^w_t \}_{t \in \mathcal{T}, d \in \{0,...,6\}}}  &\sum_{t\in \mathcal{T},d \in \mathcal{D}_t} \log \mathbb{P}(\tilde{w}_{t,d}|w_{t,d})+ \sum_{t\in \mathcal{T}}\log \mathbb{P}(g_t|p_t), \label{eq:mle obj}\\
\text{subject to: } \,
& \eqref{eq: in week 1},\eqref{eq: in week 2}, \eqref{eq: bet week 1}- \eqref{eq:bet_week_reward_belief}, \quad  t \in \mathcal{T},  d \in \{0,\cdots,6 \}, \label{eq:mle_first_const} \\
& \{c_{t,d}\}_{d=0}^{6} \in   \mathcal{C}(a_{1,t},a_{2,t},w_{t,0}, f_{b,t}, \hat{r}^w_t), \quad  t \in \mathcal{T}, \label{eq:opt in week} \\
& w_{t,d} \in \mathcal{W}, f_{t,d},c_{t,d} \in \mathcal{F}, \quad t \in \mathcal{T}, d \in \{0,...,6\}, \\
& p_t,B \in \mathcal{P}, a_{1,t},a_{2,t} \in \mathcal{A}, f_{b,t} \in \mathcal{F}, \hat{r}^w_t \in \mathcal{R}, \quad t \in \mathcal{T}.
\end{align}
\label{eq:mle_prob}
\end{subequations}
\end{proposition}%
Full details of the formulation can be found in the appendix. Here $\mathcal{T}$ is the index set of all weeks in the study and $\mathcal{D}_t$ is the set of days during week $t\in \mathcal{T}$ that have weight observations.  Note that $\mathcal{C}(a_1,a_2,w_{t,0}, f_{b,t}, \hat{r}^w_t)$ is the argmax set of \eqref{eq:in_week_model}, (i.e. is the set of decisions taken by the participant in the in-week maximization model).  \eqref{eq: in week 1} and \eqref{eq: in week 2} are first introduced in the formulation of the participant in-week decision problem and they describe the daily transitions of variables like weight ($w_{t,d}$) and  caloric intake variables ($f_{t,d}, f_{b,t}$). \eqref{eq: bet week 1}- \eqref{eq: bet week 4} correspond to the dynamics of states $a_{1,t},a_{2,t}, p_{t}, \text{and } f_{b,t}$ between consecutive weeks.

% We include the term $\prod_{(t,d) \in T\times \{1,2,...,6\}} \big(\mathbb{P}(\hat{w}_{t,d}|w_{t,d})\mathbb{P}(w_{t,d}|w_{t,d-1},c_{t,d})
%      \mathbb{P}(c_{t,d}|w_{t,d-1},a_{1,t},a_{2,t},f_{b,t})$ to deal with missing data points. Next let $\hat{w}_{t,d} = w_{t,d} + \epsilon_t$ where $\epsilon_t \sim \text{Laplace}(0,\sigma)$ and assume the set of motivational states variables are degenerate, the log-likelihood function can be simplified as follows: 
%  \begin{equation}
%      \begin{aligned}
%        \log \mathbb{P}(\{\hat{w}_{t,d}\}_{(t,d) \in T \times D},\{g_t\}_{t\in T}| \{w_{t,d}, c_{t,d}\}_{(t,d) \in T \times D }\{a_{1,t}, a_{2,t}, p_{t}, f_{b,t}\}_{t \in T}) \\
%        =\sum_{(t,d)\in T \times D} \log \mathbb{P}(\hat{w}_{t,d}|w_{t,d}) + \sum_{t\in T} \log \mathbb{P}(g_t|p_t) 
%      \end{aligned}
%  \end{equation}
Note that this formulation is non-linear and cannot be readily implemented using commercial solvers. This is due to non-linearity not only in the constraints but also in the objective function. While by assumption $\log \mathbb{P}(\tilde{w}_{t,d}|w_{t,d})$ can be expressed using linear constraints for any $t,d$, this is clearly not the case for $\log \mathbb{P}(g_t|p_t)$. Note that since $g_t \sim \text{Bernoulli}(p_t)$ we can express the p.d.f. of $g_t$ as  $\mathbb{P}(g_t|p_t) = p_t^{g_t}(1-p_t)^{1-g_t}$. So taking the log yields $\log \mathbb{P}(g_t|p_t) = g_t\log p_t + (1-g_t)\log(1-p_t)$, which is clearly non-linear and not readily expressible with mixed integer linear constraints. One approach for resolving this challenge is to use a full descritization of the objective or use a piece-wise linear approximation of the natural log function \citep{wolsey1999integer}. However, such approximations could be quite loose and have unfavorable statistical proprieties. Instead, we propose using a surrogate likelihood function \citep{bartlett2006convexity,nguyen2009surrogate,goh2018minimax,awasthi2022h}, that is a function that can be more easily deployed in commercial solvers that will produce estimators with strong statistical properties such as consistency. In particular, we choose absolute error as our surrogate for this component of the likelihood function that is $\log \mathbb{P}(g_t|p_t) \approx |p_t-g_t|$ . We will refer to the new minimization problem that only differs from \eqref{eq:mle_prob} with the substitution of $\log \mathbb{P}(g_t|p_t)$ by the surrogate function as $H_{\text{SMLE}}$. Further we will refer to a specific problem instance with observations $\{\tilde{w}_{t,d},g_t\}_{t \in \mathcal{T},d \in \mathcal{D}_t}$ and administered incentives $\{r^w_t,r^c_t\}_{t\in \mathcal{T}}$ as $H_{\text{SMLE}}( \{\tilde{w}_{t,d},g_t,r^w_t,r^c_t\}_{t \in \mathcal{T},d \in \mathcal{D}_t})$.

While using the surrogate function allows us to linearize the objective there are still two key formulation challenges in the constraints that need to be addressed so that $H_{\text{SMLE}}$ can be solved with commercial optimization software. First, $H_{\text{SMLE}}$ is a bi-level optimization problem \citep{colson2007overview,keshavarz2011imputing,bertsimas2015data,aswani2018inverse}, that is one of its constraints requires that variables be in the argmax set of a different optimization problem (usually referred to as the lower level problem). In this case, this lower level problem is a sequential decision making problem which means we will need to characterize the argmax set of an optimal policy. Second, we have nonlinear dynamics with bi-linear terms that need to be reformulated into a proper linearized form. 
% We complete the reformulation in two steps: first we solve the lower level problem $\mathcal{C}(a_{1,t},a_{2,t},w_{t,0}, f_{b,t}, \hat{r}^w_t)$ and convert it to a linear constraint in Section \ref{sec:dp soln}, then in step 2 we reformulate the set of nonlinear dynamic constraints (Equations \eqref{eq: bet week 1} - \eqref{eq: bet week 4}) as linear constraints in Section \ref{sec: mip reformulation}. 

 \subsection{Characterizing participant decisions from the in-week model}\label{sec:dp soln}

To reformulate the bi-linear constraints, we will take a direct approach by showing that  $\mathcal{C}(a_{1,t},a_{2,t},w_{t,0}, f_{b,t}, \hat{r}^w_t)$ can be characterized using a set of linear equations. To do this we will obtain a closed form solution of $c_{t,d}$ from the in-week decision model using dynamic programming. This solution is expressed in the following proposition.
\begin{proposition}
\label{prop:pat_prob_opt}
The optimal solution $\{c_{t,j}^*\}_{j=0}^6$ of the in-week decision problem is $c_{t,j}^* = f_{b,t}-\frac{a_{1,t}c\sum_{i=0}^{6-j} b^i}{2}-\frac{a_{2,t}\hat{r}^w_{t}cb^{6-j}}{4A}$ for all $j \in \{0,... 6\}$.
\end{proposition}

The complete proof can be found in Appendix \ref{sec: induction proof} here we provide a sketch. First, notice that once we take the expected value of \eqref{eq: in week 0}, then \eqref{eq:in_week_model} consists of a quadratic objective function and a set of linear constraints, meaning that this problem should have a similar optimal policy structure to a linear quadratic regulator (LQR) \citep{bertsekas2012dynamic}. This implies the value function is quadratic, and there exists a unique optimal solution for $c_{t,d}$ that can be found using backward induction, and that this solution should be a linear function of the system states. Intuitively, this optimal solution of $c_{t,d}$ matches our expectations of how financial incentives and internal motivations impact participant weight loss. To interpret these, we can consider the solution in two parts: the internal motivation component $f_{b,t}-\frac{a_{1,t}c\sum_{i=0}^{6-j} b^i}{2}$ and external motivation component $-\frac{a_{2,t}\hat{r}^w_{t}cb^{6-j}}{4A}$. The internal motivation component implies caloric intake will decrease if the internal motivation $a_{1,t}$ increases, but will otherwise be close to the participant's preferred baseline if their internal motivation is low.  The external motivation component shows that, so long as participants motivation from external compensation $a_{2,t}$ increases, so will their motivation for weight loss. Furthermore, as their expectation for future monetary incentives for weight loss increases, so too will they prefer to decrease their calories since they expect to receive a higher payoff. Interestingly, this component also shows that if participants have more uncertainty about their caloric consumption, that is $A$ increases, the effects of the financial incentives decrease. This makes intuitive sense since the more uncertainty there is in true caloric consumption, the less control participants will be able to exert on their own behavior and so receiving the reward at the end of the week becomes less certain and less motivating.

\subsection{ Reformulation of bi-linear constraints} \label{sec: mip reformulation}
Next, we reformulate the bi-linear terms in \eqref{eq: bet week 1}-\eqref{eq: bet week 3} using a set of mixed integer linear constraints and variables. First, we define two sets of binary variables $l_{1,t}, l_{2,t}$ and three sets of continuous variables $z_{1,t}, z_{2,t}, z_{3,t}$. Let $l_{1,t} \in \mathbb{B}$ be equal to 1 if the participant loses weight in week $t$, and let $l_{2,t}\in \mathbb{B}$ be equal to 1 if the probability of the participant satisfying calorie recording requirements is greater than $B$. Let $z_{1,t}$ be equal to $r^c_t\mathbbm{1}\{p_t-B\geq 0\}$ , $z_{2,t}$ be equal to $k_2 \mathbbm{1}\{(w_{\underline{d}}- w_{\bar{d}})>0$, and $z_{3,t}$ be equal to $k_{1} \mathbbm{1}\{(w_{t,0}- w_{t,6})>0\}$ . Using these quantities we will first consider the dynamics of $a_{1,t}$ from \eqref{eq: bet week 1}.
\begin{proposition}
\label{prop:reform_a_dynam}
	\eqref{eq: bet week 1} can be expressed with the following set of integer variables and constraints:
	\begin{align}
	&w_{t,0}- w_{t,6} \leq M_{1,t} (1-l_{1,t}), &  t \in \{0, \cdots, 23\}, \label{eq: a1 reform 1}\\
	&p_t-B \leq M_{2,t}l_{2,t}, &  t \in \{0, \cdots, 23\}, \label{eq: a1 reform 2}\\
	& z_{1,t} \leq M_{z1}l_{2,t}, &  t \in \{0, \cdots, 23\}, \label{eq: a1 reform 3}\\
	&z_{1,t} \leq r^c_t, & t \in \{0, \cdots, 23\}, \label{eq: a1 reform 4}\\
	&z_{1,t} \geq r^c_t - M_{z1}(1-l_{2,t}), &  t \in \{0, \cdots, 23\}, \label{eq: a1 reform 5}\\
	%& z_{1,t} \geq 0, & t \in [0, \cdots, 23] \label{eq: a1 reform 6}\\
	& z_{3,t} \leq M_{z3}l_{1,t}, &  t \in \{0, \cdots, 23\}, \label{eq: a1 reform 7}\\
	&z_{3,t} \leq k_1, &  t \in \{0, \cdots, 23\},\label{eq: a1 reform 8}\\
	&z_{3,t} \geq k_1-M_{z3}(1-l_{1,t}), &  t \in \{0, \cdots, 23\}, \label{eq: a1 reform 9}\\
	& z_{1,t},z_{3,t} \geq 0, &  t \in \{0, \cdots, 23\}, \label{eq: a1 reform 10}\\
	&a_{1,t+1} = \gamma_1(a_{1,t} - a_{1,b}) + a_{1,b} + z_{1,t} + z_{3,t}, &t\in \{0,\cdots,23\}.\label{eq: a1 reform 11}
	\end{align}

\end{proposition}

This reformulation can be done using big-M techniques for products of binary and continuous variables, as well as disjunctive constraints\citep{wolsey1999integer}, the full details can be found in the appendix.  Next we show that a similar approach can be used to reformulate the constraints that govern the dynamics of $a_{2,t}$.

\begin{proposition}
\label{prop:reform_a_2_dynam}
    \eqref{eq: bet week 2} can be expressed with the following set of integer variables and constraints:

	\begin{align}
	&w_{t,0}- w_{t,6} \leq M_{1,t} (1-l_{1,t}), &  t \in \{0, \cdots, 23\}, \label{eq: a2 reform 1}\\
	& z_{2,t} \leq M_{z2}l_{1,t}, &  t \in \{0, \cdots, 23\} ,\label{eq: a2 reform 2}\\
	&z_{2,t} \leq k_2, &  t \in \{0, \cdots, 23\}, \label{eq: a2 reform 3}\\
	&z_{2,t} \geq k2 - M_{z2}(1-l_{1,t}), &  t \in \{0, \cdots, 23\}, \label{eq: a2 reform 4}\\
	& z_{2,t} \geq 0, & t \in \{0, \cdots, 23\} \label{eq: a2 reform 5},\\
	& a_{2,t+1} = \gamma_2(a_{2,t} - a_{2,b}) + a_{2,b} + r^w_t z_{2,t}, & t\in \{0,\cdots,23\}. \label{eq: a2 reform 6}
	\end{align}

\end{proposition}

The full MILP model that incorporates these constraints and the proper surrogate objective function can be found in the appendix.  

%\begin{proposition}
%\label{prop:mle_form}
%The SMLE problem can be written as the following mixed integer program:       
%\end{proposition}

%\proof{Proof of Proposition \ref{prop:mle_form}}
%The objective function and the first 5 constraints remain the same as in the original formulation of the problem. Constraints 6-15 come directly from Proposition 1 and 16-21 come directly from Proposition 2. The last constraint ensures the set of $l_{1,t}$ and $l_{2,t}$ are binary integers. \halmos
%\endproof

%Next, we compute the posterior distribution of final weight using the mixed linear integer program. We generate a set of evenly distributed weight points centered at the estimated final weight. Then we get the log-likelihood using numerical integration. 

\subsection{Prediction and statistical consistency of surrogate likelihood estimation} \label{sec:posterior_pred_modeling}%\label{sec: consistency smle}

%add transition paragraph

While the surrogate likelihood estimation model can be thought of as descriptive, in practice clinicians are interested in predicting future participant behavior. Since we have a well defined likelihood model, we can use a Bayesian framework, similar to the one proposed in \cite{aswani2019behavioral}, in order to predict future participant behavior using this model.

To simplify the notation for this conversion, let $\theta_t = \{a_{1,t}, a_{2,t}, p_t, B, f_{b,t},\hat{r}^w_t, k_1,k_2, k_p\}$ be a shorthand for the full motivational state of the participant at week $t$, and let $\Theta = \mathcal{A}^2 \times \mathcal{P}^2 \times \mathcal{F} \times \mathcal{R}$ such that $\theta_t \in \Theta$. To convert the surrogate estimation problem into a Bayesian prediction problem we need to consider the posterior probability over the model parameters given observations $\{\tilde{w}_{t,d},g_t\}_{t\in \mathcal{T}, d \in \{0,...,6\}}$, namely $\mathbb{P}(\{\theta_t,w_{t,d},c_{t,d}\}_{t \in \mathcal{T},d \in \{0,...,6\}}| \{\tilde{w}_{t,d},g_t\}_{t\in \mathcal{T}, d \in \{0,...,6\}})$.  Using Bayes' Theorem we can write the posterior distribution in terms of the joint likelihood as follows:
\begin{multline}
    \mathbb{P}(\{\theta_t,w_{t,d},c_{t,d}\}_{t \in \mathcal{T},d \in \{0,...,6\}}| \{\tilde{w}_{t,d},g_t,r^w_t,r^c_t\}_{t\in \mathcal{T}, d \in \{0,...,6\}}) =\\ \frac{1}{Z}\mathbb{P}(\{\tilde{w}_{t,d},g_t\}_{t \in \mathcal{T}, d \in \mathcal{D}_t}| \{\theta_t, w_{t,d}, c_{t,d},r^w_t,r^c_t\}_{(t,d) \in \mathcal{T} \times \{0,...,6\} })\mathbb{P}( \{\theta_t,w_{t,d},c_{t,d}\}_{t \in \mathcal{T},d \in \{0,...,6\}}).
\end{multline}

Here, $Z$ is a normalization constant that ensures the posterior is a valid probability distribution and $\mathbb{P}( \{\theta_t,w_{t,d},c_{t,d}\}_{t \in \mathcal{T},d \in \{0,...,6\}})$ is the prior probability distribution that reflects the clinician's initial beliefs over the values of $\{\theta_t,w_{t,d},c_{t,d}\}_{t \in \mathcal{T},d \in \{0,...,6\}}$. Note that from  the structure of the model, the participant's physical and behavioral state trajectory can be fully determined if the decision maker has knowledge of the initial values of the physical and motivational states (or equivalently their current value). Thus instead of considering joint posterior and prior distributions over all possible trajectories, we will focus our formulation on distributions for the initial participant physical and motivational states $\{\theta_0,w_{0,0}\}$. Note we do not need an explicit posterior or prior on $\{c_{0,d}\}_{d=0}^6$ since by Proposition \ref{prop:pat_prob_opt} these values are fully determined by $\{\theta_0,w_{0,0}\}$. However, to obtain the posterior probability for some value of $\{\theta_0,w_{0,0}\}$ would still require us to integrate the joint posterior distribution over all possible trajectories with those initial conditions that could result in the observed data sequence $\{\tilde{w}_{t,d},g_t\}_{t\in \mathcal{T}, d \in \{0,...,6\}}$, which is numerically challenging to do. Instead, we will consider an approach similar to \cite{aswani2019behavioral} and use profile likelihood estimation to estimate the posterior distribution. For our analysis we make the following assumption on the prior distribution of $\{\theta_0,w_{0,0}\}$:
\begin{assumption}
\label{as:post_assump}
    For all $w_{0,0}\in \mathcal{W},\theta_0 \in \Theta$, $\mathbb{P}(w_{0,0},\theta_0) > 0$. Moreover, $\log\mathbb{P}(w_{0,0},\theta_0) $ can be expressed as a set of mixed integer linear constraints and objective terms. 
\end{assumption}
The first part of the assumption is key for consistency and ensures that we consider every possible value of $w_{0,0},\theta_0$ in our estimation. The second part is a relatively mild assumption that will allow us to pose the problem of obtaining our predictive estimates as a MILP. It is also satisfied by a variety of distributions such as the Laplace distribution and piece-wise linear distributions (such as those derived from histograms of previous data), of note it is also satisfied by the uniform distribution. With this in mind, consider the following optimization problem:
\begin{subequations}
\begin{align}
\eta(\bar{w}_{0,0},\bar{\theta}_0,\{r^w_t,r^c_t\}_{t\in \mathcal{T}} )= \notag \\
\min_{\{w_{t,d}, \theta_{t},c_{t,d}\}_{t \in \mathcal T, d \in \{0,...,6\}}}&  \sum_{t\in \mathcal{T},d \in \mathcal{D}_t} -\log \mathbb{P}(\tilde{w}_{t,d}|w_{t,d})+ \sum_{t\in \mathcal{T}}|g_t-p_t| - \log \mathbb{P}(w_{0,0},\theta_{0,0}) +\log Z, \label{eq:post_obj}\\
\text{subject to: } \,
& \eqref{eq: in week 1},\eqref{eq: in week 2}, \eqref{eq: bet week 1}- \eqref{eq:bet_week_reward_belief}, \quad  t \in \mathcal{T},  d \in \{0,\cdots,6\},\\
&\{c_{t,d}\}_{d=0}^{6} \in   \mathcal{C}(a_{1,t},a_{2,t},w_{t,0}, f_{b,t}, \hat{r}^w_t), \quad  t \in \mathcal{T},\label{eq:post_opt in week} \\
& w_{0,0} = \bar{w}_{0,0}, \theta_{0,0} = \bar{\theta}_{0}, \label{eq:init_cond_post} \\
& w_{t,d} \in \mathcal{W}, f_{t,d},c_{t,d} \in \mathcal{F}, \quad t \in \mathcal{T}, d \in \{0,...,6\}, \\
& p_t,B \in \mathcal{P}, a_{1,t},a_{2,t} \in \mathcal{A}, f_{b,t}\in \mathcal{F}, \hat{r}^w_t \in \mathcal{R}, \quad t \in \mathcal{T}.
\end{align}
\label{eq:post_prob}
\end{subequations}
Note that \eqref{eq:post_prob} is essentially the same formulation as $H_\textbf{SMLE}$ with the addition of the log prior and normalization terms to the objective and Constraint \eqref{eq:init_cond_post} that sets the initial conditions. Problem \eqref{eq:post_prob} is in fact a feasibility problem, that when solved evaluates a function $\eta: \mathcal{W}\times \Theta\times\mathcal{R}^{2|\mathcal{T}|} \mapsto \mathbb{R}$, which is very similar to the log posterior distribution, but uses the surrogate likelihood instead of the true joint likelihood. By removing \eqref{eq:init_cond_post} and the term $\log Z$ from the objective, we can transform \eqref{eq:post_prob} into a problem that calculates the surrogate maximum a posteriori estimate (MAP) for $w_{0,0},\theta_0$, we will call these estimates $\hat{w}^\text{MAP}_{0,0},\hat{\theta}^\text{MAP}_{0}$. One challenge with \eqref{eq:post_prob} is that the value of $Z$ is not generally known and must be estimated by solving \eqref{eq:post_prob} at several initial conditions and then using numerical integration. Alternatively, we can estimate a surrogate posterior using the MAP estimates at a particular value of $\bar{w}_{0,0},\bar{\theta}_0$ as follows:
\begin{equation}
    \hat{\mathbb{P}}(\bar{w}_{0,0}, \bar{\theta}_0| \{\tilde{w}_{t,d},g_t,r^w_t,r^c_t\}_{t\in \mathcal{T}, d \in \{0,...,6\}}) = \frac{\exp(-\eta(\bar{w}_{0,0}, \bar{\theta}_0, \{r^w_t,r^c_t\}_{t\in \mathcal{T}}) )}{\exp(-\eta(\hat{w}^\text{MAP}_{0,0}, \hat{\theta}^\text{MAP}_0, \{r^w_t,r^c_t\}_{t\in \mathcal{T}}))} \label{eq:sur_post_probab}
\end{equation}
Using this posterior distribution we can characterize the uncertainty around the initial conditions and form predictions and scenarios for future participant behavior.

\subsection{Consistency proof}
\label{sec:consist_proof}
We now proceed to prove that the estimates computed by $H_{\text{SMLE}}$ and the predictive model are statistically consistent, that is, as more data is collected from the participant these estimates become closer to their ground truth value (or a value that is closest to the true distribution given the model definition). This condition is key for ensuring that any adaptive framework that is using a stream of participant data can provide effective incentives that are properly personalized to each participant. Moreover, this is a necessary condition to ensure such an adaptive policy is asymptotically optimal. In general, proving surrogate likelihood functions yield consistent estimates requires that the estimation problem have Lipschitz continuous objective function and constraints (and by extension a Lipschitz continuous value function) that allows using known asymptotic and finite time bounds \citep{bartlett2006convexity,nguyen2009surrogate}. Since our estimation problem is a MILP, we do not necessarily satisfy this continuity condition. On the other hand, analysis of consistency of MILP based parameter estimates relies on an exact optimal solution of the optimization problem with respect to the true joint likelihood function of the problem \citep{mintz2017behavioral}. Clearly, in the case of surrogate likelihood estimation this condition is not satisfied and so a different analysis is required. Our approach will extend the results for consistency of MILP estimates to the case of surrogate estimation, when the surrogate loss is within a multiplicative constant of the true likelihood. While we focus our analysis on the participant model in the context of weight loss interventions, the technique presented here can be generalized to any surrogate estimation using MIPs with a bounded likelihood function. 

%Next, we prove the posterior estimates from SMLE model is consistent, which implies our estimates will converge to the ground truth as $t \rightarrow \infty$. Statistical consistency is formally defined as follows,

%To simplify the notation for our analysis, let $\theta_t = \{a_{1,t}, a_{2,t}, p_t, B, f_{b,t},\hat{r}^w_t\}$ be a shorthand for the full motivational state of the participant at week $t$, and let $\Theta = \mathcal{A}^2 \times \mathcal{P}^2 \times \mathcal{F} \times \mathcal{R}$ such that $\theta_t \in \Theta$.
Let $\{\hat{w}_{0,0},\hat{\theta}_0\} \in \argmin H_{\text{SMLE}}( \{\tilde{w}_{t,d},g_t,r_t^w,r^c_t\}_{t \in \mathcal{T},d \in \mathcal{D}_t})$ be the estimates calculated in the surrogate likelihood estimation problem, and let $w^*_{0,0}, \theta^*_0$ be the true value of these parameters for a particular participant. To show that $\hat{w}_{0,0},\hat{\theta}_0 \overset{p}{\to} w^*_{0,0}, \theta^*_0 $ we will first show that the surrogate posterior probability function defined in \eqref{eq:sur_post_probab} is consistent in the Bayesian sense, which would then imply that $\hat{w}^\text{MAP}_{0,0},\hat{\theta}^\text{MAP}_{0}$ are consistent estimators for any prior distribution that satisfies Assumption \ref{as:post_assump}. Because the uniform distribution satisfies this assumption, and because under a uniform prior $\hat{w}^\text{MAP}_{0,0},\hat{\theta}^\text{MAP}_{0} = \hat{w}_{0,},\hat{\theta}_0$ this would mean that the $H_\text{SMLE}$ estimates are also consistent. To formally conduct our analysis we will need the following definition for Bayesian consistency of a posterior distribution:

% \subsubsection{Bayesian Prediction}
% To convert the surrogate estimation problem into a Bayesian prediction problem we need to consider the posterior probability over the model parameters given observations $\{\tilde{w}_{t,d},g_t\}_{t\in \mathcal{T}, d \in [0,...,6]}$, namely $\mathbb{P}(\{\theta_t,w_{t,d},c_{t,d}\}_{t \in \mathcal{T},d \in [0,...,6]}| \{\tilde{w}_{t,d},g_t\}_{t\in \mathcal{T}, d \in [0,...,6]})$. 

\begin{definition}
\label{def:bayes_consist}
For all $(w_{0,0}^*, \theta_0^*) \in \mathcal{W} \times \Theta$ and constants $r, \delta >0$, we say the estimate of the posterior distribution $\hat{\mathbb{P}}(\cdot|\{\tilde{w}_{t,d},g_t,r^w_t,r^c_t\}_{t\in \mathcal{T}, d \in \{0,...,6\}})$ is consistent if $\mathbb{P}_{(w_{0,0}^*, g_0^*, \theta_0^*)} (\hat{\mathbb{P}}(S(\delta)|\{\tilde{w}_{t,d},g_t,r^w_t,r^c_t\}_{t\in \mathcal{T}, d \in \{0,...,6\}}) \geq r) \rightarrow 0 \text{ as } t \rightarrow \infty $. Here $\mathbb{P}_{(w_{0,0}^*, \theta_0^*)}$ is the probability law where $(w_{0,0}^*, \theta_0^*)$ are the true initial conditions of the system, and  where
$S(\delta):=\{(w_{0,0}, \theta_0) \not\in \mathcal{B}((w_{0,0}^*, \theta_0^*),\delta)\}$, where $\mathcal{B}((w_{0,0}^*, \theta_0^*),\delta)$ is an open ball with radius $\delta$ centered around $(w_{0,0}^*, \theta_0^*)$.
\end{definition}

The implication of Definition \ref{def:bayes_consist} is that if our posterior estimate is consistent, then as more data is collected it turns into a degenerate distribution at the true parameter values. While this is a stronger condition then parameter consistency we will show our estimate possesses this property and that this implies the point estimates are consistent as well. To proceed with the analysis we make the following technical assumption.

\begin{assumption} \label{assumption: p_t}
There exists  $\epsilon > 0$ such that the set $\mathcal{P} := [\epsilon, 1-\epsilon]$. In other words, for all $t \in \mathcal{T}$, $\epsilon\leq p_t \leq 1-\epsilon$.
\end{assumption}

This assumption ensures that $\mathcal{P}$ is a compact set making it easily deployable with commercial optimization solvers. It also ensures that the value of $p_t$ and by extension $\mathbb{P}(g_t|p_t)$ is bounded, which will be key in showing that surrogate posterior estimates are consistent. In practice this is a reasonable assumption since it guarantees that on any week in the trial a participant will have some positive probability of successfully completing their calorie recording goal or failing it. This is reflected in real-world interventions where no participant truly has an almost sure probability of failing to record or recording their calories. We will also require the following assumption on the history of the observations.

%The probability $p_t$ is bounded between 0 and 1 to reflect the real world scenario that no participant is guaranteed to always fail or succeed in a weight loss intervention (cite). Computationally, this assumption also allows us to bound the log-likelihood of $|g_t-p_t|$, which is crucial for proving the statistical consistency. 

\begin{assumption}
\label{as:sufficient_excite}
Let $(w_0^*, \theta_0^*)$ be the true initial conditions, the incentives $\{r^w_t, r^c_t\}_{t \in \mathcal{T}}$ are such that for any $\delta > 0$,
\begin{equation}
    \max_{S(\delta)} \lim_{|\mathcal{T}| \rightarrow \infty} \sum_{t \in \mathcal{T},d \in \mathcal{D}_t} -\log \frac{\mathbb{P}(\tilde{w}_{t,d}|\bar{w}_{t,d})}{\mathbb{P}(\tilde{w}_{t,d}|w_{t,d})} + \sum_{t \in \mathcal{T}}- \log \frac{\mathbb{P}(g_t|\bar{p}_t)}{\mathbb{P}(g_t|p_t)}= -\infty.
    \end{equation}
    where $\bar{w}_{t,d},\bar{p}_t$ are the states and decisions under initial conditions $(w_{0,0}, p_{0}) \in S(\delta)$, and $w_{t,d}, p_t$ are the states and decisions under true initial conditions $(w_{0,0}^*, p_{0}^*)$.
\end{assumption}
This assumption is known as a sufficient excitation condition and is a common assumption in the literature \citep{craig1987adaptive,aastrom2013adaptive}. Essentially, this assumption states that there is sufficient variance from the incentives administered so that it is possible for the clinician to identify the true states of the participants. In practice this assumption can be satisfied if there is sufficient process noise or by adding random perturbations to the incentives administered. Using this assumption, we can now proceed to prove the consistency of the posterior estimate. First, we prove a proposition on the structure of the surrogate likelihood function. 
\begin{proposition}
\label{prop:surogate}
  Given Assumptions \ref{as:post_assump}-- \ref{as:sufficient_excite}, $|g_t-p_t|$ can be bounded as:
  \begin{equation}
      \frac{-\log(1-\epsilon)}{\epsilon}|g_t-p_t|\leq-\log(\mathbb{P}(g_t|p_t))\leq \frac{-\log(\epsilon)}{1-\epsilon}|g_t-p_t|. \label{eq:smle_mle_bound}
  \end{equation}
\end{proposition}

The complete proof can be found in the appendix, and here we present a brief sketch. Using Assumption \ref{assumption: p_t}, we consider two cases (one when $g_t=0$ and one when $g_t = 1$) and use a calculus argument to show that the desired bounds hold. From \eqref{eq:smle_mle_bound}, we see that so long as $p_t$ is bounded then $\log(\mathbb{P}(g_t|p_t)) = \Theta(|g_t - p_t|)$, this will be key in showing convergence since it implies these expressions have similar asymptotic behavior. We note that the keys to this proposition are that the probability measure is log concave and bounded. Without these conditions, there could be edge-case observations that would make it difficult to distinguish between underlying values of $p_t$. With this structure we can now prove the main result on the posterior estimate.
%
%Given Assumption \ref{assumption: p_t}, we show each bound is either monotonically increasing or monotonically decreasing and  details of the proof can be found in Appendix \ref{app:surogate}.

\begin{proposition}
\label{prop:consistency}
Given Assumptions \ref{as:post_assump}-- \ref{as:sufficient_excite}, the surrogate posterior estimate $\hat{\mathbb{P}}(w_{0,0}, \theta_0| \{\tilde{w}_{t,d},g_t,r^w_t,r^c_t\}_{t\in \mathcal{T}, d \in \{0,...,6\}})$ is consistent.
\end{proposition}
The complete proof of this proposition can be found in the appendix here we present a sketch. The main arguments are first to use Proposition \ref{prop:surogate} to create a point-wise upper bound for the surrogate posterior function in terms of the true posterior function specified by the model. Then by Assumption \ref{as:sufficient_excite} we show that this bound implies that for any initial conditions $(w_{0,0},\theta_0) \neq (w^*_{0,0},\theta^*_0)$ the posterior assigns zero probability in the limit. To complete the proof we use a volume bound to show that this condition holds uniformly over $\mathcal{W} \times \Theta$. This proposition shows that our posterior estimates satisfy Definition \ref{def:bayes_consist}, and implies the following corrollary.
\begin{corollary}
\label{cor:consist}
Given Assumptions \ref{as:post_assump}-- \ref{as:sufficient_excite}, $(\hat{w}^\text{MAP}_{0,0},\hat{\theta}_0^\text{MAP})\overset{p}{\to}(w^*_{0,0},\theta_0^*) $.
\end{corollary}
%We prove the consistency using the bounds found in Proposition \ref{prop:surogate} and the sufficient excitation assumption. Details of the proof can be found in Appendix \ref{app:consistency}.
%\subsection{Use For Bayesian Prediction}
The complete proof of the corollary can be found in the appendix. Note that since Corollary \ref{cor:consist} holds for surrogate maximum \textit{a posteriori} estimates calculated with any prior distribution that satisfies Assumption \ref{as:post_assump}, including the uniform distribution. However, if we use the uniform prior, then our predictive problem is exactly $H_\text{SMLE}$ meaning that the estimators calculated from this problem are also consistent.

\section{Financial incentive optimization }\label{sec:alg}
In this section, we show how the model and prediction techniques from Section \ref{sec: mle} can be used to optimize personalized financial incentives for a cohort of participants in a weight loss trial. Recall that the goal of the interventionist is to administer financial incentives to each participant to maximize the number of participants that achieve clinically significant weight loss by the end of the trial while remaining within the intervention budget. Furthermore the incentive administered should seem random to the participant. To formally define our problem, let $U$ be the set of participants. For this section, we will augment the notation from Sections \ref{sec: mle} and \ref{sec:pat_prob} by including an additional index of $u \in U$ to indicate parameters specific to a trial participant $u$. So, for instance, the weight and motivational states of participant $u$ at week $t$ and day $d$ will be given by $w_{u,t,d},\theta_{u,t}$ respectively. Let $\mathcal{L}:\mathcal{W} \to \mathbb{R}$ be a loss function that captures if a participant is unable to lose a clinically significant amount of weight. We leave this loss function in a general form since there are several ways of designing this incentive optimization problem depending on the interventionist's secondary outcomes, we present some illustrative examples of loss functions in Section \ref{sec:opt_incentive_design}. In week $t$, the clinician calculates a distribution $\pi_{u,t} \in \Delta_{\mathcal{R}^2}$ for each participant $u \in U$, where $\Delta_{\mathcal{R}^2}$ is the set of distribution with support over $\mathcal{R}^2$, and administers incentive $\{r^w_{u,t}, r^c_{u,t}\} \sim \pi_{u,t}$. Let $G$ be the total intervention budget, that is, the clinician requires that with probability one $\sum_{u \in U, t \in \mathcal{T}} r^w_{u,t} + r^c_{u,t} \leq G$. The ultimate goal of the clinician is to find a sequence of distributions for all participants $\{\pi_{u,t}\}_{u\in U,t \in \mathcal{T}}$ such that $\mathbb{E}\sum_{u \in U}\mathcal{L}(w_{u,24,6}(\{\pi_{u,t}\}_{t \in \mathcal{T}}))$ is minimized and the budget constraint is not violated, where the expectation is taken over not only the uncertainty in the participant parameter values but also over the stochasticity of the incentive distribution. 

As stated in this general form, this problem is challenging to solve due to the presence of a hard constraint, partially observed parameters, and randomized policy. Moreover, using standard techniques such as scenario generation \citep{kaut2003evaluation} may not be tractable since different scenarios need to be created not only for each potential value of the unobserved states but also for each realization of the reward distribution and each participant. Instead we will consider a different approach that leverages the statistical properties of our posterior estimates from Section \ref{sec: mle} and certainty equivalence to approximate a solution to the interventionist's problem. Specifically, we propose an adaptive approximation approach, where at each time $t$ the interventionist will estimate the unknown participant parameters using $H_{SMLE}$ for each participant and then calculate an incentive design based on these estimates. For our approximation approach, we restrict our policies to be only the set of deterministic policies over $\mathcal{R}$, equivalently distribution $\pi_{u,t}$ where for each $u\in U$ they assign a probability mass of 1 to a single element of $\mathcal{R}$. This restriction will simplify our formulation since we will not need to consider different realizations of the incentive distribution and can concentrate our efforts on the uncertainty in the unobserved participant parameters. Moreover, it will ensure that we can easily meet the budget constraint with probability one. In practice, financial incentives are rarely truly random in weight loss interventions but are in fact predetermined by interventionists to be perceived as random by participants \citep{leahey2015benefits,almeida2015effectiveness}. Since our adaptive approximation approach will be recomputing incentives at each time period, despite using a deterministic policy, since these rewards will be frequently changing, they should still be perceived as random by study participants making this approach suitable for our setting. In the remainder of this section, we will first present the details of our approximation algorithm and then provide guarantees that our method is asymptotically optimal over the class of deterministic policies. This guarantee ensures that under proper technical conditions the policy calculated by our method will converge to the best deterministic policy as more data is collected form the participants over the course of the intervention.

\subsection{Approximation algorithm for personalized incentive design} \label{sec: dia}
To form our adaptive approach, we will consider a framework where interventionists minimize their loss with respect to their posterior information at each time step. To formalize this, suppose that it is currently the start of week $T$ (where $1<T<24$) of the intervention, then let $\mathcal{F}_T = \{\tilde{w}_{u,t,d}, g_{u,t},r_{u,t}^w, r_{u,t}^c\}_{u\in U, t \in \{0,...,T\},d\in \mathcal{D}_t}$. Our approach will solve the following deterministic policy problem formulation.
\begin{align}
   \min_{\{r_{u,i}^w, r_{u,i}^c\}_{i=T}^{24} \in \mathcal{R}^2}  \{\mathbb{E}[\sum_{u\in U}\mathcal{L}(w_{u,24,6})|\mathcal{F}_T] : \sum_{u\in U,t\in \mathcal{T}} r_{u,t}^w+ r_{u,t}^c \leq G \}.\label{eq:posterior_loss_min}
\end{align}

From the modeling assumptions in Section \ref{sec: mle} we note that knowledge of $(w_{u,0,0},\theta_{u,0})$ for each participant are sufficient to determine the trajectory of all other parameters for participant $u$ given a  sequence of incentives. Thus by the smoothing theorem \citep{bickel2015mathematical}, there exists some function $\phi:\mathcal{W}\times\Theta \times \mathcal{R}^2 \to \mathbb{R}$ such that \eqref{eq:posterior_loss_min} can be reformulated as:
\begin{align} \label{eq: id real problem}
   \min_{\{r_{u,i}^w, r_{u,i}^c\}_{i=T}^{24} \in \mathcal{R}^2} \{ \mathbb{E}[\sum_{u \in U} \phi(w_{u,0,0},\theta_{u,0},\{r_{u,t}^w, r_{u,t}^c\}_{t \in \mathcal{T}})|\mathcal{F}_T]: \sum_{u\in U,t\in \mathcal{T}} r_{u,t}^w+ r_{u,t}^c \leq G \}. 
\end{align}

In general the closed form of $\phi$ is difficult to obtain since it relies on the composition of the loss function and model dynamics; however, this reformulation illustrates that in order to approximate the expectation in the objective we would only need to consider an estimate of the posterior distribution for $(w_{u,0,0},\theta_{u,0})$, such as the posterior estimate in \eqref{eq:sur_post_probab}. Thus one approach for solving \eqref{eq: id real problem} is using scenario generation and discretizing $\mathcal{W}\times \Theta$ into a grid of $m$ scenarios. This would result in the following optimization problem:

\begin{subequations}
\begin{align}
\min_{\{r_{u,i}^w, r_{u,i}^c\}_{i=T}^{24}} & \sum_{u \in U,k = \{0,...,m\}} \phi(w^k_{u,0,0},\theta^k_{u,0},\{r_{u,t}^w, r_{u,t}^c\}_{t \in \mathcal{T}}) \hat{\mathbb{P}}(w^k_{u,0,0}, \theta^k_{u,0}| \mathcal{F}_T), \\
\text{subject to: }& \sum_{u\in U,t\in \mathcal{T}} r_{u,t}^w+ r_{u,t}^c \leq G, \\
& r_{u,t}^w,r_{u,t}^c \in \mathcal{R}^2.
\end{align}
\end{subequations}

Solving this optimization problem is challenging first because the set $\Theta$ is high dimensional meaning that a large number of grid points may need to be selected in order to obtain a sufficiently close approximation to the distribution. Furthermore, recall that to compute $\hat{\mathbb{P}}(w^k_{u,0,0}, \theta^k_{u,0}| \{\tilde{w}_{u,t,d},g_{u,t},r^w_{u,t},r^c_{u,t}\}_{t\in \mathcal{T}, d \in \{0,...,6\}})$ requires solving a MIP for each $k \in \{0,..m\}$. Thus to form the objective would require solving $|U|m$ MIPs, which can be computationally expensive and would be challenging to scale to large weight loss interventions. Instead of using a full posterior approach, we instead propose to use the either the surrogate MAP or MLE estimates of $w_{u,0,0}, \theta_{u,0}$ with data up to time $T$, that we will denote as $\hat{w}^T_{u,0,0}, \hat{\theta}^T_{u,0}$, as single point estimates and optimizing future incentives with respect to these estimates. We formalize this problem as follows:
\begin{subequations}   
\label{eq:incentive_design_problem}
\begin{align}
  \psi_T (\{\hat{w}^T_{u,0,0},\hat{\theta}^T_{u,0},\{\bar{r}^w_{u,t}, \bar{r}^c_{u,t}\}_{t=0}^{T}\}_{u\in U})
    = &\min \sum_{u\in U}\mathcal{L}(w_{u,24,6}) ,\label{eq: dia 1}\\
   \text{subject to:}  &\sum_{u \in U} \sum_{t \in W} r^w_{u,t} + r^c_{u,t}  \leq G ,\label{eq: dia 2}\\
   &\text{Constraints \eqref{eq:mle_first_const}-\eqref{eq:opt in week}}, \quad \forall u\in U, \forall t \in \{0,...,24\}, \label{eq: dia 3} \\
   & w_{u,0,0} = \hat{w}^T_{u,0,0}, \hat{\theta}^T_{u,0} = \theta_{u,0}, \quad \forall u \in U, \\ &
   r^w_{u,t} = \bar{r}^w_{u,t}, r^c_{u,t} =  \bar{r}^c_{u,t} \quad \forall u \in U, \forall t = \{0,...,T\}, \\
   & w_{u,t,d} \in \mathcal{W}, \theta_{u,t} \in \Theta, \quad \forall u \in U, t \in \{0,...,24\}.
\end{align}
\end{subequations}
Here the values $\bar{r}^w_{u,t}, \bar{r}^c_{u,t}$ are the previously administered financial rewards from the beginning of the intervention up to the current time period $T$. Therefore $\psi_T$ should be interpreted as the minimum possible value of the loss function if the true initial conditions of each participant $u$ are the estimates $\{\hat{w}^T_{u,0,0},\hat{\theta}^T_{u,0}\}$ and the rewards that they have received up to the current time period are fixed to their historical values.

\begin{algorithm}
\caption{Design of Incentives Algorithm (DIA)}\label{alg:incentive}
\begin{algorithmic}
\State \Require $\{\tilde{w}_{u,t,d},g_{u,t},r_{u,t}^w,r^c_{u,t}\}_{t \in \mathcal{T},d \in \mathcal{D}_{u,t}}$ for all $u \in U$,
\State Compute $(\{\hat{w}^T_{u,0,0},\hat{\theta}^T_{u,0}\}_{u \in U}) \in \argmin H_{\text{SMLE}}(\{\tilde{w}_{u,t,d},g_{u,t},r_{u,t}^w,r^c_{u,t}\}_{t \in \mathcal{T},d \in \mathcal{D}_{u,t}})$,
\State Compute $\{r^w_{u,t}, r^c_{u,t}\}_{t\in \{T,...,24\}, u \in U} \in \argmin\{ \psi_T (\{\hat{w}^T_{u,0,0},\hat{\theta}^T_{u,0},\{\bar{r}^w_{u,t}, \bar{r}^c_{u,t}\}_{t=0}^{T}\}_{u\in U})| \{r^w_{u,t}, r^c_{u,t}\}_{t=T}^{24})\}$,
\State Apply $r^w_{u,T}, r^c_{u,T}$ back to $u \in U$.
\end{algorithmic}
\end{algorithm}

Using this formulation we define our adaptive incentive calculation approach that we call the Design of Incentives Algorithm (DIA).  The pseudocode of DIA is presented in Algorithm \ref{alg:incentive}, and consists of three main steps. First, all data up to the current time step $T$ is used to estimate model parameter for each participant using the SMLE model ($H_{\text{SMLE}}$) established in Section \ref{sec: mle}. Then, using the parameter estimates of all participants and previously dispensed incentives as inputs, we solve \eqref{eq:incentive_design_problem} to compute a sequence of incentives from period $T$ to 24. We then apply the incentive values for period $T$, $\{r_{u,T}^w,r_{u,T}^c\}$ to each participant $u \in U$ and collect new observations. These three steps are repeated for each week until we reach the end of the intervention. As new data is collected the parameter estimators are updated and new incentives are computed. 

% As more data is collected over time, our proposed algorithm provides better incentive design that help minimize a selected loss function. We formulate the step of optimizing incentive using parameter estimates as a incentive design model, in which the incentive design minimizes a loss function under a defined budget.

%  We formulate the incentive design model in Equations \ref{eq: dia 0}-\ref{eq: dia 3}. We define $U$ as the set of participants and $G$ as the budget of the intervention. Next, let $u$ be the index of participant contained in the participant set $U$, $w$ be the week index, and $d$ be the day index. Then denote $w_{u,w,d}$ as the weight of participant $u$ on day $d$ of week $w$, and $a_{u,1,w}, a_{u,2,w}$ as participant $u$'s motivational states for weight loss and weight loss financial incentive in week $w$. Equation \ref{eq: dia 1} is the general loss function, where $\mathcal{W}_u$ denotes the weight trajectory of participant $u \in U$ and $g_u \in \mathcal{B}$ indicates whether or not the calorie recording requirement is satisfied. Since in the real life settings, limited budget for financial incentives and cost-effectiveness of the intervention are major concern in the field of study, we include Constraint to ensure the total financial incentive distributed to the set of participants spent never exceeds a budget $B$. The additional set of constraints is the same as the constraints in $H_{\text{SMLE}}$ which captures the dynamics of weight-related parameters. 

\subsection{Asymptotic optimality} \label{sec: asymptotic opt}
Here we show that the incentives output by DIA are asymptotically optimal with respect to the class of deterministic policies. This property ensures that as more data is collected from each participant over the course of the intervention, DIA produces incentives that approach the optimal incentives with respect to a full information problem, with policies restricted to the set of deterministic policies. In this section, we present sketches of proofs of each proposition and the detailed proofs can be found in the appendix.

Our proof approach will be similar to that proposed by \cite{mintz2017behavioral} with modification to our setting. In general, asymptotic optimality is not trivial to guarantee since it requires that the optima of an approximation problem converge in probability to the optima of the goal problem being approximated. Point-wise convergence of the value functions is usually insufficient to prove this property, and often it requires uniform convergence of the value function of the approximation problems to the objective of the goal problem. However, since our approximations are based on MIP formulations proving uniform convergence maybe difficult to guarantee. Thus, we will use a weaker condition known as epi-convergence \citep{lachout2005strong} that is sufficient to prove this result. This condition ensures that the epigraph of the value functions of the approximation problems converges stochastically to the epigraph of the target problem, and thus ensures convergence of the lower-level sets and minima.

For our analysis we will need to define the following value function:

\begin{subequations}
\label{eq:incentive_feasibility}
    \begin{align}
  \psi (\{\bar{w}_{u,0,0},\bar{\theta}_{u,0},\{\bar{r}^w_{u,t}, \bar{r}^c_{u,t}\}_{t=0}^{24}\}_{u\in U})
    = &\min \sum_{u\in U}\mathcal{L}(w_{u,24,6}) ,\label{eq: dia 1}\\
   \text{subject to:}  &\sum_{u \in U} \sum_{t \in W} r^w_{u,t} + r^c_{u,t}  \leq G, \label{eq: dia 2}\\
   &\text{Constraints \eqref{eq:mle_first_const}-\eqref{eq:opt in week}}, \quad \forall u\in U, \forall t \in \{0,...,24\}, \label{eq: dia 3} \\
   & w_{u,0,0} = \bar{w}_{u,0,0}, \bar{\theta}_{u,0} = \theta_{u,0}, \quad \forall u \in U,\\ &
   r^w_{u,t} = \bar{r}^w_{u,t}, r^c_{u,t} =  \bar{r}^c_{u,t}, \quad \forall u \in U, \forall t = \{0,...,24\}, \\
   & w_{u,t,d} \in \mathcal{W}, \theta_{u,t} \in \Theta, \quad \forall u \in U, t \in \{0,...,24\}.
\end{align}
\end{subequations}
Note that $\psi (\{\bar{w}_{u,0,0},\bar{\theta}_{u,0},\{\bar{r}^w_{u,t}, \bar{r}^c_{u,t}\}_{t=0}^{24}\}_{u\in U})$ is the value function of a problem quite similar to \eqref{eq:incentive_design_problem}. However, unlike \eqref{eq:incentive_design_problem}, \eqref{eq:incentive_feasibility} is a feasibility problem where the incentive sequence is predefined for the entirety of the intervention and not only up to time $T$. Thus  $\psi (\{\bar{w}_{u,0,0},\bar{\theta}_{u,0},\{\bar{r}^w_{u,t}, \bar{r}^c_{u,t}\}_{t=0}^{24}\}_{u\in U})$ can be interpreted as the minimum loss that would be expected by offering the predetermined sequence $\{\bar{r}^w_{u,t}, \bar{r}^c_{u,t}\}_{t=0}^{24}$ to each participant if their individual parameter values where truly equal to $\{\bar{w}_{u,0,0},\bar{\theta}_{u,0}\}$. To begin, our analysis we will show that $\psi (\{\bar{w}_{u,0,0},\bar{\theta}_{u,0},\{\bar{r}^w_{u,t}, \bar{r}^c_{u,t}\}_{t=0}^{24}\}_{u\in U})$ has structural properties that insure that given our estimation is consistent will result in asymptotically optimal incentives.

\begin{proposition} \label{prop: semicont}
If Assumptions \ref{as:post_assump}-\ref{as:sufficient_excite} hold, then the value function $ \psi(\{w_{u,0,0},\theta_{u,0},\{r^w_{u,t}, r^c_{u,t}\}_{t=0}^{T+n}\}_{u \in U})$ is lower semi-continuous in each argument $\{w_{u,0,0}\}_{u \in U}$, $\{\theta_{u,0}\}_{u \in U}$, and $\{\{r^w_{u,t}, r^c_{u,t}\}_{w=T+1}^{T+n}\}_{u\in U}$. 
\end{proposition}

To prove this proposition we first show the problem can be reformulated as a parametric MILP with each of the parameter arguments as affine terms in the constraints, and then apply the results from \cite{hassanzadeh2014generalization}. This proposition ensures that the value function $ \psi(\{w_{u,0,0},\theta_{u,0},\{r^w_{u,t}, r^c_{u,t}\}_{t=0}^{T+n}\}_{u \in U})$ has a closed epigraph and closed lower level sets, a key property for showing the convergence of minima.

For the remainder of the analysis, let $w^*_{u,0,0},\theta^*_{u,0}$ be the true initial parameter values for each participant $u \in U$ and as before let $\{\hat{w}^T_{u,0,0},\hat{\theta}^T_{u,0}\}$ be the estimates provided by $H_\text{SMLE}$ estimates of these parameters at time $T$.  Using the structure from Proposition \ref{prop: semicont} we analyze the manner by which $ \psi (\{\hat{w}^T_{u,0,0},\hat{\theta}^T_{u,0},\{\bar{r}^w_{u,t}, \bar{r}^c_{u,t}\}_{t=0}^{24}\}_{u\in U})$ converges to $ \psi (\{w^*_{u,0,0},\theta^*_{u,0},\{\bar{r}^w_{u,t}, \bar{r}^c_{u,t}\}_{t=0}^{24}\}_{u\in U})$. In particular we show the following convergence property.

\begin{proposition}\label{prop: semicont approx}
If Assumptions \ref{as:post_assump}-\ref{as:sufficient_excite}  hold, then $ \psi (\{\hat{w}^T_{u,0,0},\hat{\theta}^T_{u,0},\{\bar{r}^w_{u,t}, \bar{r}^c_{u,t}\}_{t=0}^{24}\}_{u\in U}) \xrightarrow[\mathcal{R}^{2|U|})]{l-prob} \psi (\{w^*_{u,0,0},\theta^*_{u,0},\{\bar{r}^w_{u,t}, \bar{r}^c_{u,t}\}_{t=0}^{24}\}_{u\in U})$, 
which means the function $ \psi (\{\hat{w}^T_{u,0,0},\hat{\theta}^T_{u,0},\{\bar{r}^w_{u,t}, \bar{r}^c_{u,t}\}_{t=0}^{24}\}_{u\in U}) $ is a lower semi-continuous approximation to the function $\psi (\{w^*_{u,0,0},\theta^*_{u,0},\{\bar{r}^w_{u,t}, \bar{r}^c_{u,t}\}_{t=0}^{24}\}_{u\in U})$ \cite{lachout2005strong}.
\end{proposition}

to prove this proposition we apply Proposition \ref{prop:consistency} and Proposition \ref{prop: semicont} in conjunction with results from \cite{lachout2005strong}. This property ensures that any lower level set centered around some incentive sequence $\{\{\bar{r}^w_{u,t}, \bar{r}^c_{u,t}\}_{t=0}^{24}\}_{u\in U}$ for the value function evaluated at the estimates, will converge in probability to the lower level set of the corresponding problem with the initial parameters equal to their true values. Note that this is a stronger structural property then simple point-wise convergence since this condition must hold for any lower level set of $\psi$ on the incentive space $\mathcal{R}^{2|U|}$. This property also essentially ensures that the value functions of the sequence of approximation problems that use the $H_\text{SMLE}$ estimates will converge to the epigraphs of the value function of the problem with the true parameter values.  This property leads us to the final result that shows the solution provided by DIA is asymptotically optimal for the participant's true initial conditions. 

\begin{theorem}\label{thm:asymp opt}
Denote the set of optimal deterministic financial incentives under the true initial conditions $\{(w^*_{u,0,0},\theta^*_{u,0})\}_{u\in U}$ as $R^*(\{(w^*_{u,0,0},\theta^*_{u,0})\}_{u\in U}) := \argmin\{\psi (\{w^*_{u,0,0},\theta^*_{u,0},\{\bar{r}^w_{u,t}, \bar{r}^c_{u,t}\}_{t=0}^{24}\}_{u\in U})|\{\{\bar{r}^w_{u,t}, \bar{r}^c_{u,t}\}_{t=0}^{24}\}_{u\in U} \}\}$.  If Assumptions \ref{as:post_assump}-\ref{as:sufficient_excite}  hold and $dist(x,Y)=\inf_{y \in Y}||x-y||$, then 
\begin{equation}
\begin{aligned}
    dist&\big(\{r^{w,DIA}_{u,T}, r^{c,DIA}_{u,T}\}_{u \in U},
    R^*(\{(w^*_{u,0,0},\theta^*_{u,0})\}_{u\in U})\}\big) \xrightarrow{p} 0 
\end{aligned}
\end{equation}
for any $\{r^{w,DIA}_{u,T}, r^{c,DIA}_{u,T}\}_{u \in U}$ returned by Algorithm DIA as $T \rightarrow \infty$. 
\end{theorem}

We prove the final results by combining the results of Proposition \ref{prop:consistency}, \ref{prop: semicont}, and \ref{prop: semicont approx}. This result implies that as additional data is collected by the clinician on the participants, the recommended incentives calculated by DIA will approach the optimal deterministic incentives that should be allocated to each participant. The two keys to this result are that estimates computed from $H_\text{SMLE}$ are consistent and that our problem structure results in lower semi-continuous value functions. Note that while this result shows asymptotic optimality with respect to the class of deterministic policies, it does not provide guarantees on how DIA would fair against the best stochastic policies, an analysis that is more complex to conduct analytically. In Section \ref{sec:experiment}, we  provide an empirical examination of several stochastic policies and compare their performance to DIA.
%\section{Consistency Analysis}
%\[\min |g(x)-\hat{g}(x)|\]
%\[ \min \sum_{w,d}|w_{w,d}-\hat{w}_{w,d}|+\sum_{w}|p_w - g_w|\]

%\begin{itemize}
  
    %\item $a_{1,w+1} = \gamma_1(a_{1,w} - a_{1,b}) + a_{1,b} + k_{1} \mathbbm{1}\{(w_{\underline{d}}- w_{\bar{d}})>0\} +  r^w_w\mathbbm{1}\{p_w-B\geq 0\}
   
   % \item $a_{2,w+1} = \gamma_2(a_{2,w} - a_{2,b}) + a_{2,b} + k_2 r^w_w \mathbbm{1}\{(w_{\underline{d}}- w_{\bar{d}})>0 \}$
    %\item $a_{3,w+1}= \gamma_3(a_{3,w} - a_{3,b}) + a_{3,b} - k_3\hat{r}_{c,w}g_w$
   % \item $a_{4,w+1}= a_{4,w}=\beta$
   % \item $p_{w+1}=\gamma_p(p_w-p_{w,b})+p_{w,b}+k_pg_w $
%\end{itemize}
%$f: \mathbb{R}^n \mapsto \mathbb{B}$ such that $g_{w+1} = f(g_w, \hat{r}_{w+1c}, r_{wc}, a_{1,w},a_{2,w},a_{3,w},f_{b,w})$

\section{Numerical studies}\label{sec:experiment}

We conducted three sets of numerical studies using data from the Log2Lose trial \citep{voils2018study}. The first study analyzed the performance of our methodology for capturing a participant's true weight trajectory. For this study 
we fit the SMLE model to the weight records of participants with different weight trajectories for the entire 24 weeks of the trial and show how well our predicted trajectory fits this data. The second study examined the accuracy of our predictive method in predicting a participant's weight trajectory using weight and incentive data from a short time span. We compared the predictive performance of our behavioral model against three machine learning methods (logistic regression, linear support vector machine (SVM), and random forest).The third study examine how our DIA method performs in designing financial incentives to maximize clinical weight loss successes.   For this study we compare the efficacy of different financial incentive policies (deterministic, randomized, one-size-fits-all in Log2Lose) under different budget options using DIA in terms of number of participants able to achieve clinically significant weight loss and percentage of weight lost by the five participants who lost the least weight. 

Our results show our approach is well-suited for capturing different weight trajectories and predicting the future trajectory. In terms of the financial incentives design, our results show that the deterministic and randomized policies, where the incentives are generated by DIA, are more effective for encouraging weight loss than the one-size-fits-all policy implemented in the original Log2Lose study. In addition, the results show the randomized policy is potentially better suited for weight loss intervention than the deterministic policy. We ran all the experiments in Python \citep{10.5555/1593511} and compute the optimization problems using Gurobi v9.1.1 \citep{gurobi}.

%At current stage we only explore the performance of 75\% and 25\% randomized policy, Given the observations that showed the participant's adherence to the intervention improved more in most cases with randomized policy, we plan to focus on studying the optimal randomized policy based on personal behaviors and different time duration (i.e. short-term weight loss or maintaining weight loss in the long run). All experiments are run on a MacBook (intel core i7 processor and 16GB memory). The optimization models are computed using Gurobi v9.1.1.

\subsection{Describing different weight loss trajectories}
\label{sec:ddwt}
\begin{figure}
     \centering
     \begin{subfigure}[b]{0.4\textwidth}
         \centering
         \includegraphics[width=\textwidth]{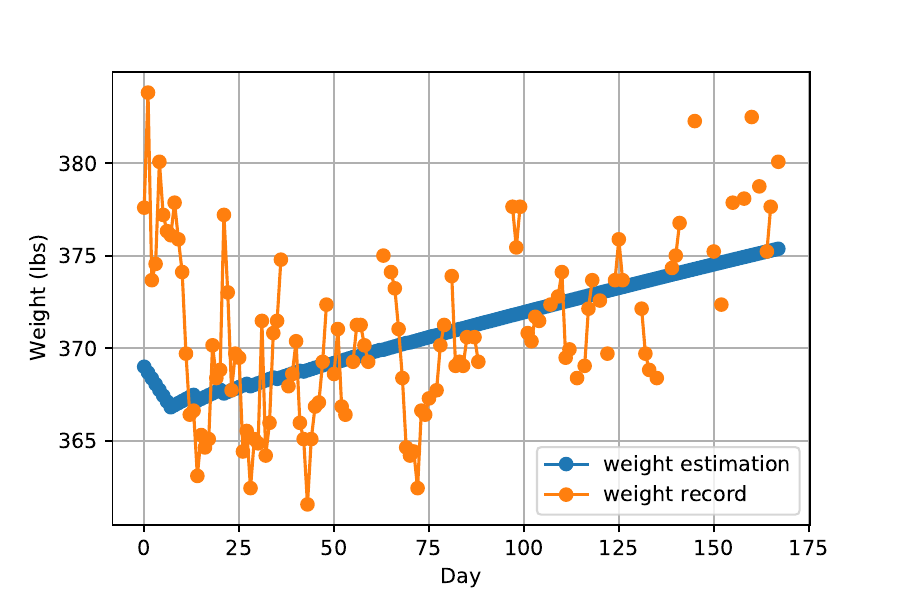}
         \caption{}
         \label{fig:fit trajectory up}
     \end{subfigure}
     %\hfill
     \begin{subfigure}[b]{0.4\textwidth}
         \centering
         \includegraphics[width=\textwidth]{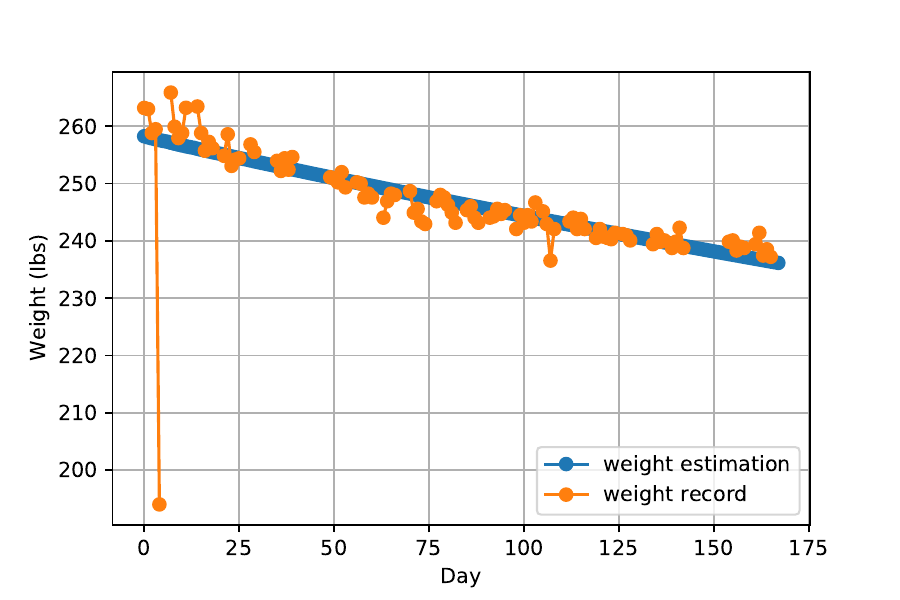}
          \caption{}
         \label{fig:fit trajectory down}
     \end{subfigure}
     
     \begin{subfigure}[b]{0.4\textwidth}
         \centering
         \includegraphics[width=\textwidth]{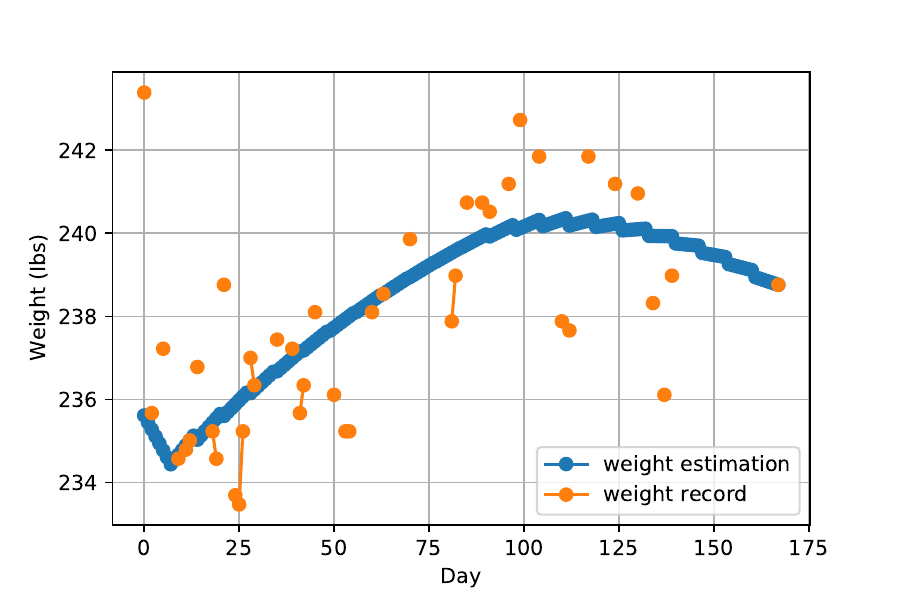}
          \caption{}
         \label{fig:fit trajectory curve}
     \end{subfigure}
     \begin{subfigure}[b]{0.4\textwidth}
         \centering
         \includegraphics[width=\textwidth]{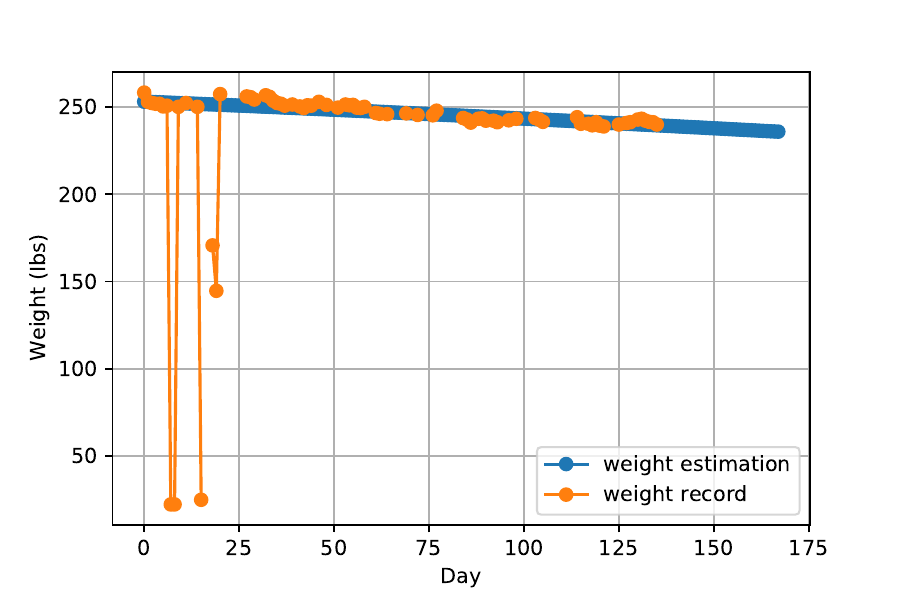}
          \caption{}
         \label{fig:fit trajectory error}
        
     \end{subfigure}
        \caption{4 examples of comparisons of true weight trajectory (orange) and the estimated fitting weight trajectory (blue) for week 0-24. }
        \label{fig:fit trajectory}
\end{figure}

In this study, we examine how well a behavioral model trained with $H_\text{SMLE}$ is able to capture different weight loss trajectories using the entire 24 weeks of data from the Log2Lose trial. We found the weight loss trajectories fit three common patterns: 1) participants who lose weight initially but then later become resistant to the intervention, 2) participants who lose weight consistently over the course of the intervention, and 3) participants who are resistant to the intervention and do not lose much weight. We name these groups initial achievers, constant achievers, and intervention-resistant, respectively. and note that out of 67 participants they make up 10\%, 73\%, and 5\% of study participants, respectively. The remaining 12\% of study participants had too few weight and calorie records for this analysis and were thus excluded. %All other participants that did not fit neatly into these patterns we labeled as abnormal.

The results in Figure \ref{fig:fit trajectory} show that using our behavioral model, the estimated trajectory is a good fit to the observed trajectory regardless of the missing weight records or weight loss pattern. Figure \ref{fig:fit trajectory up} shows how our model fits an early achiever, Figure \ref{fig:fit trajectory down}  shows the fit to a constant achiever, and Figure \ref{fig:fit trajectory curve}  shows the fit to a trajectory of an intervention resistant participant. Figure \ref{fig:fit trajectory error} shows the predicted weight trajectory remains accurate even when a participant's weight records have multiple missing consecutive measurements and anomalous measurements. These anomalous weight measurements could be caused by another household member of the study participant (or a pet) stepping onto the cellular scale. This result indicates that our proposed method is insensitive to these measures and can extract the underlying weight loss trajectory of the study participant.

%The difficult level of satisfying the requirements for these two incentives are quite different. Users can simply weigh themselves by standing on a scale provided by the clinicians, and the weight is automatically recorded. However, it requires much more effort to satisfy the dietary incentive. 

%\begin{figure}
    %\centering
    %\includegraphics[width=0.6\textwidth]{w.png}
   % \caption{Fit Weight Loss Trajectory: The x-axis is the index of days in the program and y-axis is the weight (lbs). The orange scatter plot represents the true weights and blue scatter plot represents the estimated weights generated by the predictive model}
   % \label{fig:galaxy}
%\end{figure}

 \subsection{Comparison of predictive performance}
In this numerical study, we examined the performance of our behavioral model to predict whether or not a participant achieves clinically significant weight loss, defined as at least 5\% weight loss, at the end the study (week 24).  We compare our model against three common machine learning methods: linear SVM, logistic regression, and random forest \citep{breiman2001random,hastie2009elements}. For this prediction task we generated the labels by selecting either the weight at the end of the program or the last available weight record of week 24 (if the final weight is missing) as the true final weight of the participant and setting it to 1 if the final weight was no more than 95\% of the initial weight and zero otherwise. For this study we included data from a total of 67 study participants who had at least 1 weight record in week 24.

Since our behavioral model performs a regression task, we used our posterior estimate from Section \ref{sec:posterior_pred_modeling} to compute the probability the final weight would be below the clinically significant level using numerical integration (in a manner similar to \cite{aswani2019behavioral}). Then we varied a prediction threshold such that if this probability was larger than threshold our model would predict a label of 1. Our behavioral model only used daily weight and calorie measures and weekly incentive amounts as data for prediction.

All three machine learning methods were implemented using scikit-learn \citep{scikit-learn}. For these models we used age, gender, height, body mass index, weekly average weight, two types of financial incentives, and weekly average caloric intake as training features. Since our data contained missing daily records, we could not directly use the data records. As an alternative option, we used weekly averages of caloric intake and weight since most weeks contained at least some measures of these features. We evaluated the predictive performance of these methods using five-fold cross validation, where in each fold 80\% of the participants were used as a training set and 20\% were used for validation. Within each fold we used another round of five-fold cross validation to optimize the hyperparameters of each of these ML methods. 
%
% For the participants in the test set of each fold we performed each prediction task at  
%
% We split the data set by putting 80\% of participants' data into a training set and 20\% of participants to the testing set. Once we split the participants, we ran the study with feature sets that captured different time spans of study weeks.

\begin{figure}
    \centering
    \includegraphics[width=0.3\textwidth]{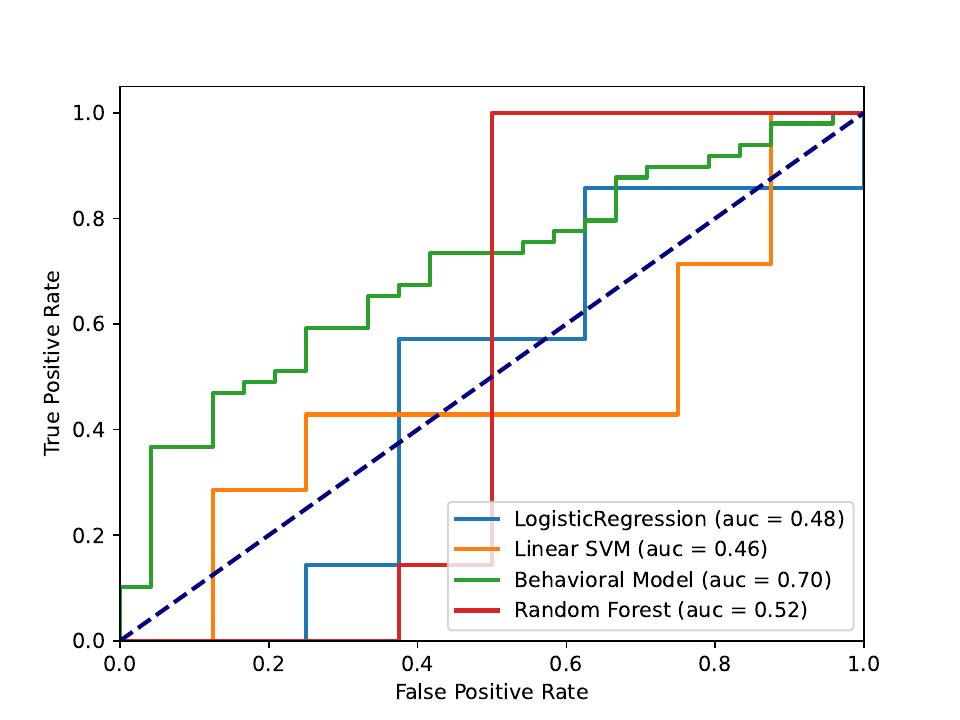}
    \includegraphics[width=0.3\textwidth]{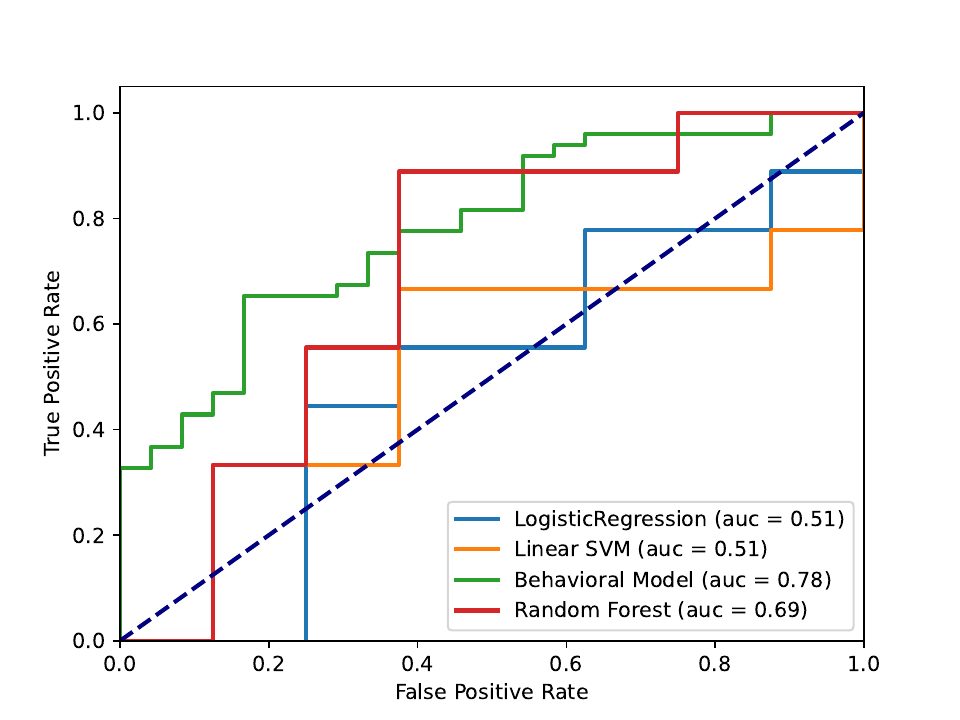}\\
    \includegraphics[width=0.3\textwidth]{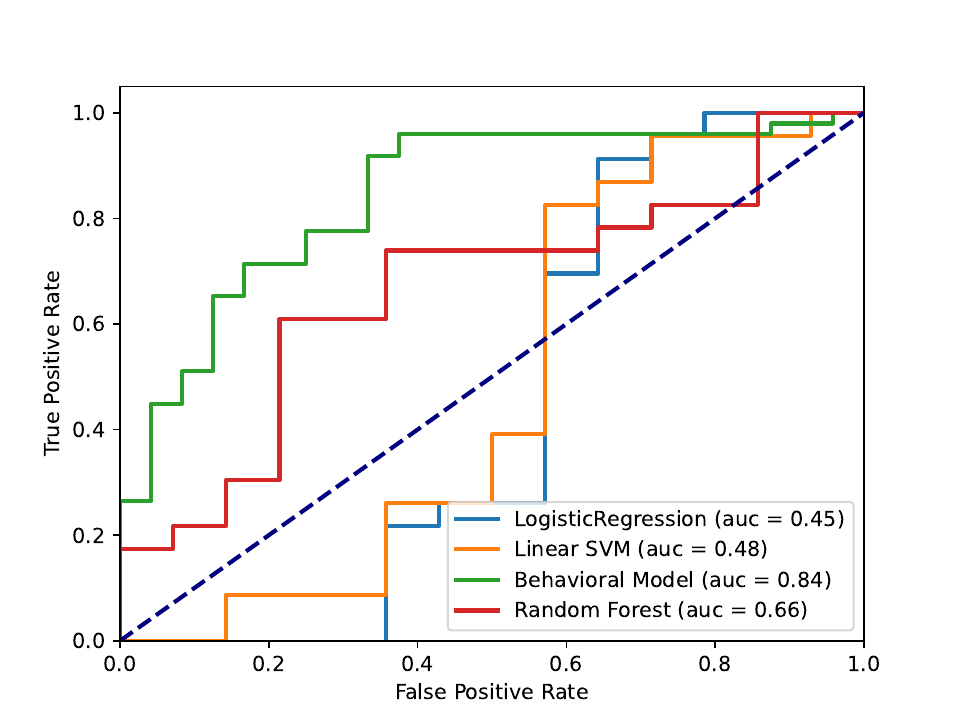}
    \includegraphics[width=0.3\textwidth]{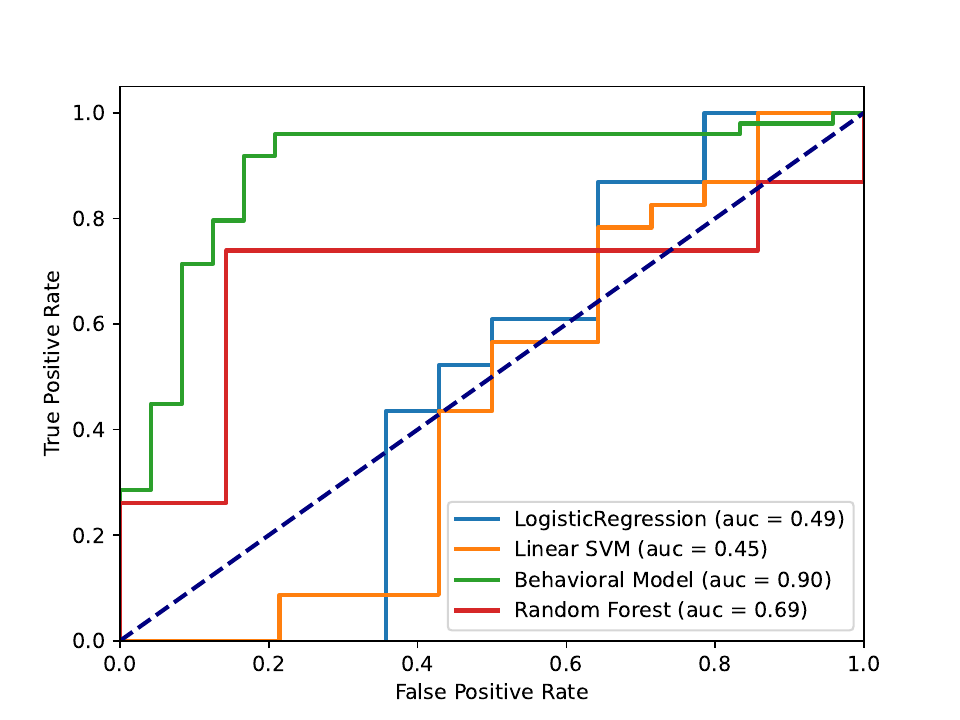}
    
    \caption{Raw ROC curves for various number of training weeks: (top: 4 weeks(left), 8 weeks(right); below: 16 weeks(left), 20 weeks(right).}
     \label{fig:raw_roc}
\end{figure}

To see how well each model is capable of using limited data and avoid over fitting we fit each model with feature sets that captured the first 4, 8, 12, 16, and 20 weeks. Note that, for each setting the models were tasked with predicting weight loss by week 24, meaning models trained on 4 weeks of data were predicting a measure 20 weeks in the future, models with 8 weeks of training data were predicting weight loss 16 weeks in the future, and so forth.  We computed the false and true positive rates of each model and plotted them as ROC curves to analyze their predictive performance.  Figure \ref{fig:raw_roc} show the raw ROC curves for each time span. The figure shows that the performance of logistic regression and linear SVM does not improve as data from additional weeks is incorporated. Using random forest, we observe moderate improvement until the number of training weeks reaches 20. In contrast, our behavioral model improves consistently in its predictive capability as additional data is incorporated from the study participants. This suggests our proposed method is better suited for weight loss prediction even in the early weeks of a weight loss intervention.  Our results also validate the consistency of the parameter estimates computed using $H_\text{SMLE}$. The results show our proposed behavioral model performs significantly better than the ML methods for longer training weeks and outperforms the ML methods for shorter training weeks with low false positive rate (FPR $\leq 0.4$). For instance, using the first 16 weeks of data as the training set, with a false positive rate at 0.43, the highest true positive rate achieved by the machine learning methods is 0.74 (random forest) while the true positive rate of our model is 0.92.  We also note that with 20 weeks of data, the only competitive method to our behavioral model is the random forest predictor; however, the highest true positive it can achieve is 0.74 with a false positive between 0.21 and 0.78, and it is never able to achieve the 0.92 true positive rate which our model is able to achieve with a false positive of 0.17. This indicates that the random forest model is likely over fitting to data and may not be appropriate for incentive optimization in this setting, while our method is capable of leveraging the participant data effectively for prediction and optimization.
% \begin{figure}
%     \centering
%     \includegraphics[width=0.3\textwidth]{smoothroc_4n.pdf}
%     \includegraphics[width=0.3\textwidth]{smoothroc_8n.pdf}\\
%     \includegraphics[width=0.3\textwidth]{smoothroc_16n.pdf}
%     \includegraphics[width=0.3\textwidth]{smoothroc_20n.pdf}
    
%     \caption{Smooth ROC curves for various number of training weeks: (top: 4 weeks(left), 8 weeks(right); below: 16 weeks(left), 20 weeks(right).}
%      \label{fig:smooth_roc}
    
% \end{figure}

\subsection{Simulation study of optimal incentive design}
\label{sec:opt_incentive_design}
In the third study, we examine how well our adaptive methods perform in a simulated weight loss trial, and how deterministic policies compare to stochastic ones. We examine seven different incentive policies and examine their performance in terms of the number of participants able to achieve clinically significant weight loss, and the amount of weigh lost by the five participants who lost the least amount of weight.
%
%implemented the DIA, and compared the efficacy of four incentive policies and two types of loss functions. 
The policies we examined included six optimization based incentive policies that varied in whether they considered stochastic or deterministic incentives and what loss function they were optimizing. We also evaluated the incentive schedule implemented by the study investigators of Log2Lose. %optimization based policies one deterministic and two stochastic) each evaluated with two different loss functions resulting  computed by DIA , a 75\% randomized policy, and 25\% randomized policy and the Log2Lose policy. 
The deterministic optimization policies where based off of our proposed DIA method using a parameter estimate computed with $H_\text{SMLE}$, and distributed exactly the incentive amount computed by this method to each participant. To test stochastic policies that are still able to satisfy the budget constraint with probability one, we considered policies that at each week $T$ would either provide a participant with their incentive amount as computed by DIA for some loss function or zero incentive with some non-zero probability $q \in \{0.25,0.75\}$.
%When 75\% randomized policy is implemented, each participant receives the optimal incentive with a probability of 0.75. Similarly, with 25\% randomized policy each participant receives the optimal incentive with a probability of 0.25. Notice the weekly optimal incentive is the optimal solution based on the estimated parameters, and it is not necessarily the true optimal incentive design.
For these optimization policies we considered two different loss functions:  the indicator loss ($\min \sum_{u \in \mathcal{U}} \mathbbm{1}\{w_{u,23,6} \leq 0.95w_{u,0,0}\}$) and the hinge loss ($\min \sum_{u \in \mathcal{U}} (w_{u,23,6}-0.95w_{u,0,0})^+$). The indicator loss minimizes the number of participants who lose less than 5\%   of weight, and the hinge loss minimizes the gap between the final weight and the 5 \% percent weight loss goal for participants who did not meet the weight goal.

We only included those study participants who had sufficient data and participated in the three treatment arms (A,B,C) of Log2Lose and were thus eligible for financial incentives. This resulted in the data of 47 participants being included in this study. %The two loss functions are the indicator loss ($\min \sum_{u \in \mathcal{U}} \mathbbm{1}\{w_{u,23,6} \leq 0.95w_{u,0,0}\}$) and the hinge loss ($\min \sum_{u \in \mathcal{U}} (w_{u,23,6}-0.95w_{u,0,0})^+$). The indicator loss minimizes the number of participants who lose less than 5 percent of weight, and the hinge loss minimizes the gap between the final weight and the 5 percent weight loss goal for participants who did not meet the weight goal.
%
%We ran the policy comparison task following the same simulation process detailed in the proposed algorithm. First we implemented the original Log2Lose policy and used estimated parameters of the first 2 weeks to start the recursion.
%
For each participant, we fit our behavioral model on their full study data (much like in Section \ref{sec:ddwt}) using $H_\text{SMLE}$, and used these fitted dynamics to simulate their behavior over the course of the trial. In each simulated week the particular incentive computation method would use available weight and recording goal measurements to compute a set of incentives for all 47 participants. Then the participants would receive this incentive and their dynamics would advance with the same functional and noise structure as detailed in Section \ref{sec:pat_prob}. To simulate the noise over the trial we generate each new measurement of $g_{t,u}$ from a Bernoulli distribution with a mean equal to their respective $p_{t,u}$, set the value of $A = 500$ to reflect uncertainty in caloric intake of being within 500 calories, and set the variance of the Laplace noise of $w_{u,t}$ with a variance of $8$ (parameter $b=2$) derived from the empirical variance of weight measures observed in our data. Each simulated trial was run with 5 replicates. To ensure our estimation methods had sufficient observations to provide parameter estimates, we initialized each simulated trial with a two week run-in period where incentives were allocated at the same values they were disbursed in the Log2Lose trial. Thus from the second incentive given to each participant on-wards, our optimization based methods began to differ from the incentives given by Log2Lose. To test how effective each method is with respect to intervention budget we ran simulated trials with 10 budget options in the range of \$520-\$5,857. The reason our range starts at \$520  is because this is the amount of money distributed to the group of participants in week 1 of the trial by Log2Lose (and thus during our simulated run-in period). We then constructed our range by increasing the budget by \$100 increments until we reached \$920. Since \$920 is approximately 15\% of the total amount of incentives disbursed during Log2Lose, the remaining budgets we examined were at 20\%, 40\%, 60\%, 80\%, and 100\% of the total amount spent which was \$5,857. When the budget was set to \$520, participants received no financial incentives after the first week regardless of the policy choice. We note that, because each participant received the same incentive as in the Log2Lose study the performance of the Log2Lose policy could not be evaluated at different budget levels other then what was observed in the data.

\begin{figure}
     \centering
     \begin{subfigure}[b]{0.4\textwidth}
         \centering
         \includegraphics[width=\textwidth]{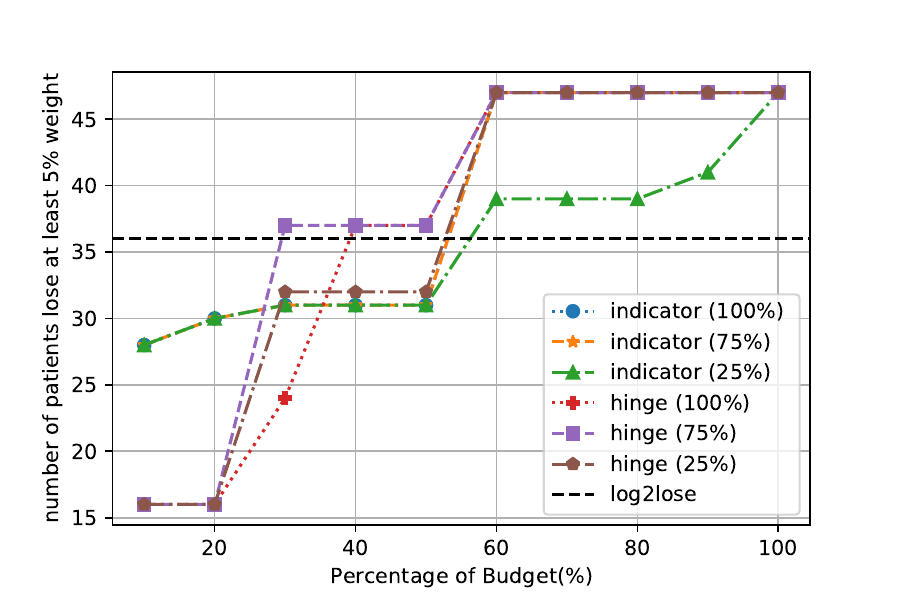}
         \caption{}
         \label{fig:budget_patient}
     \end{subfigure}
     %\hfill
     \begin{subfigure}[b]{0.4\textwidth}
         \centering
         \includegraphics[width=\textwidth]{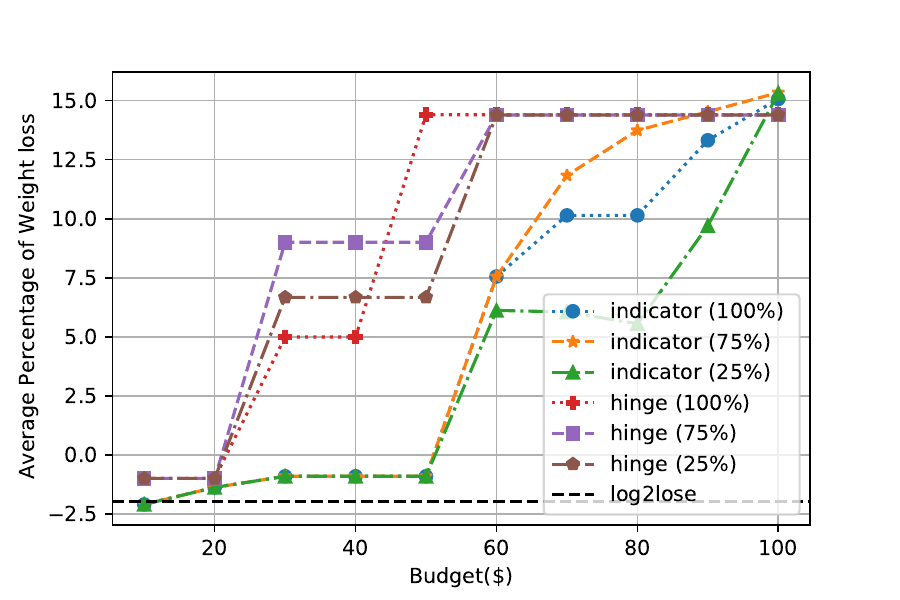}
          \caption{}
         \label{fig:budget_percent}
     \end{subfigure}
    
        \caption{Number of participant achieving clinical weight loss success ( $\geq$ 5\% weight loss) (\ref{fig:budget_patient}) and average percentage of weight loss across the bottom 5 participants who lose the least weight (\ref{fig:budget_percent}) by the end of week 24   using 6 incentive policies and the randomized policy implemented in the Log2lose trial.}
        \label{fig:budget_combined}
\end{figure}

Figure \ref{fig:budget_patient} shows a comparison of the number of participants who achieved at least 5\% weight loss with incentives provided by each of the different policies at different budget levels. Each optimization based approach is labeled as either indicator or hinge depending on the loss function used in its optimization; the percentage corresponds to the probability of the participant receiving the DIA incentive (with 100\% corresponding to the DIA method). From this figure, we can see that our methods are able to achieve comparable performance to the Log2Lose policy with 20-60\% of the budget spent during the Log2Lose trial. Moreover, from our simulation results, all optimization policies are able to assist nearly the whole participant cohort in achieving clinically significant weight loss when using 100\% of the budget used by Log2Lose. This indicates that through our optimization-based approach, and predictive modeling, we are able to allocate incentives to participants when they are most likely to assist them in weight loss. Furthermore, since our approaches are personalized and not one-size-fits all, they are able to provide participants who are more externally motivated with greater incentives amounts to promote weight loss. This is in contrast to the one-size-fits-all approach, that is restricted in providing the same incentive schedule to all participants and thus spends some part of the budget on participants who may not need the added incentive to promote weight loss. Interestingly, the policy that is capable of achieving performance comparable to Log2Lose with the least amount of budget is a policy that only provides the DIA incentive with 75\% probability and uses the hinge loss and not the deterministic DIA policy with the indicator loss. This indicates that by making the incentives intermittent-an approach consistent with psychological learning theory- weight loss behaviour can be promoted effectively and potentially more efficiently.

%
%When the objective of the budget model is the indicator function, the DIA and  75\% randomized policies ensure all 47 participant lose at least 5 percent of their initial weight by the end of the intervention with a budget greater than or equal to \$1070. Even though the 25\% randomized optimal reward policy is not as effective as the 75\% randomized policy, it still outperforms the Log2Lose policy. Using the hinge loss objective function, the results show the randomized policies outperform the Log2Lose and deterministic policies when the budget is at least \$920. 

% \begin{figure}
%     \centering
%     \includegraphics[width=0.4\textwidth]{budget_numPatient.pdf}
 
%     \caption{Number of participant achieving clinical weight loss success ( $\geq$ 5\% weight loss) by the end of week 24 using 6 incentive policies and the randomized policy implemented in the Log2lose trial.}
%      \label{fig:budget_patient}
% \end{figure}

Figure \ref{fig:budget_percent} shows the average percentage of weight loss achieved by the five participants who lost the least percentage of weight over the 24 weeks. The results show both deterministic and randomized policies outperform the Log2Lose policy for a wide range of budgets, again reaffirming that, through personalization, resources can be spent on participants who are more likely to respond to financial incentives and thus promote overall weight loss. Although the deterministic DIA policy guarantees each participant receives incentives and should prioritize weight loss by all participants with the hinge loss objective the randomized policies outperform the deterministic policies. Again, this reaffirms the effectiveness of intermittent incentives, and suggests that, in practice, a form of randomized policy could be effective in implementation.

% \begin{figure}
%     \centering
%     \includegraphics[width=0.4\textwidth]{budget_percentLoss.pdf}
 
%     \caption{Average percentage of weight loss across the bottom 5 participants who lose the least weight at the end of week 24 using 6 incentive policies and the randomized policy implemented in the Log2lose trial.}
%      \label{fig:budget_percent}
% \end{figure}

%Next, we compared the ratio of the motivational states $a_1, a_2$ at the end of the Log2Lose trial with 100\%, 80\%, or 60\% of the budget. When implementing the hinge loss objective function, the average ratio of $a_1$ and $a_2$ is not affected by the change in budget. When the objective function is the indicator function, the average ratio of $a_1$ and $a_2$ decreases as the budget increases, which suggests that as the budget decreases, each patient is less likely to receive incentive in the latter weeks of the intervention. However, a moderate reduction in budget did not affect the performance of the algorithm. As in the case of taking the hinge loss objective, we see no change in the log ratio of $a_1$ and $a_2$. \ym{need some more discussion here about what this means from a behavioral perspective}

%\begin{figure}
   % \centering
   % \includegraphics[width=0.4\textwidth]{loga1a2Ratio.png}
 
    %\caption{Log of the average ratio of a1 to a2 in week 24 across different budgets}
    % \label{fig:Loga1a2}
%\end{figure}

\begin{figure}
     \centering
     \begin{subfigure}[b]{0.4\textwidth}
         \centering
         \includegraphics[width=\textwidth]{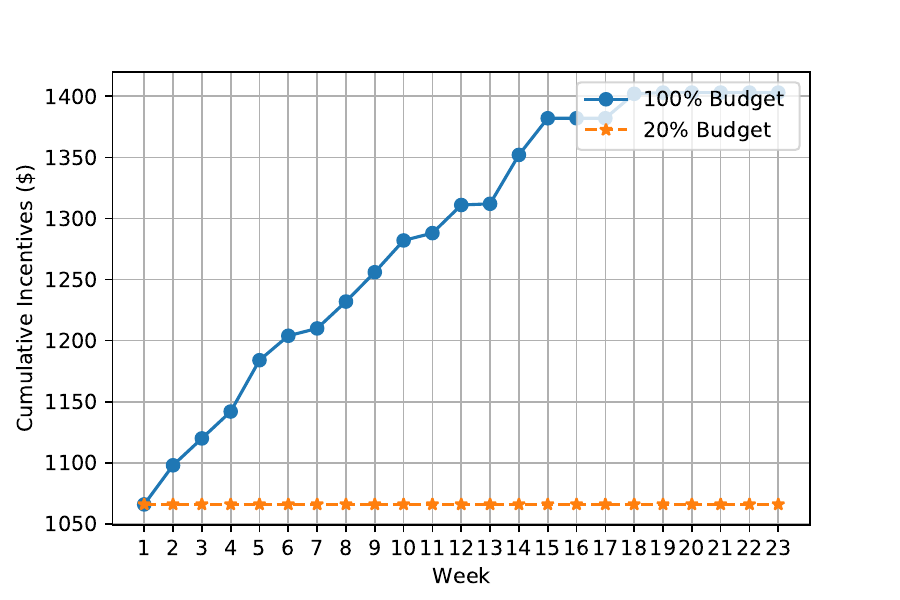}
         \caption{}
         \label{fig:hinge cumulative}
     \end{subfigure}
     %\hfill
     \begin{subfigure}[b]{0.4\textwidth}
         \centering
         \includegraphics[width=\textwidth]{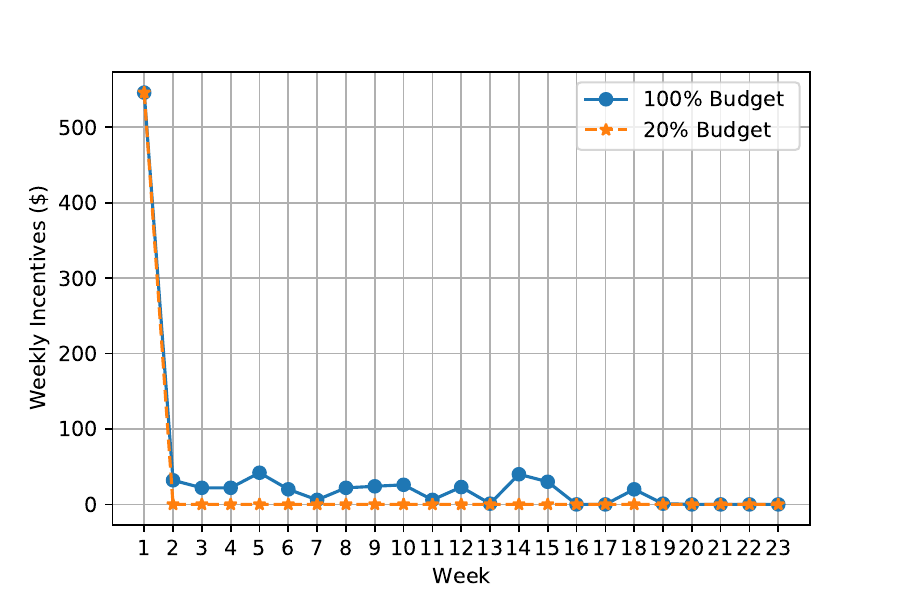}
          \caption{}
         \label{fig:hinge marginal}
     \end{subfigure}
    
        \caption{Implementing the hinge loss function and deterministic incentive policy, Figure \ref{fig:hinge cumulative} shows the cumulative incentives distributed with 100\% and 20\% budget and Figure \ref{fig:hinge marginal} shows the incentives distributed per week with 100\% and 20\% budget.}
        \label{fig:early large incentives}
\end{figure}

\begin{figure}
     \centering
     \begin{subfigure}[b]{0.4\textwidth}
         \centering
         \includegraphics[width=\textwidth]{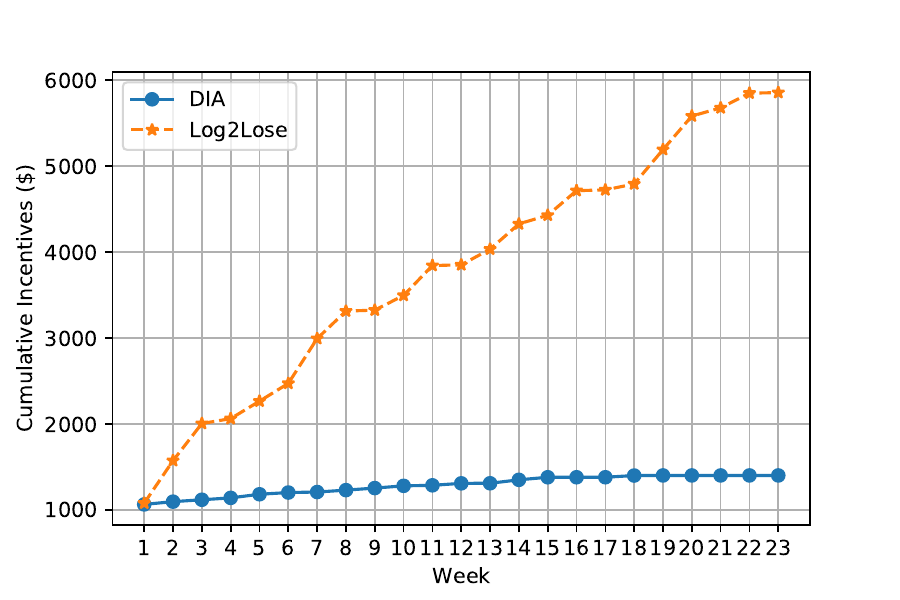}
         \caption{}
         \label{fig:cumulative budget}
     \end{subfigure}
     %\hfill
     \begin{subfigure}[b]{0.4\textwidth}
         \centering
         \includegraphics[width=\textwidth]{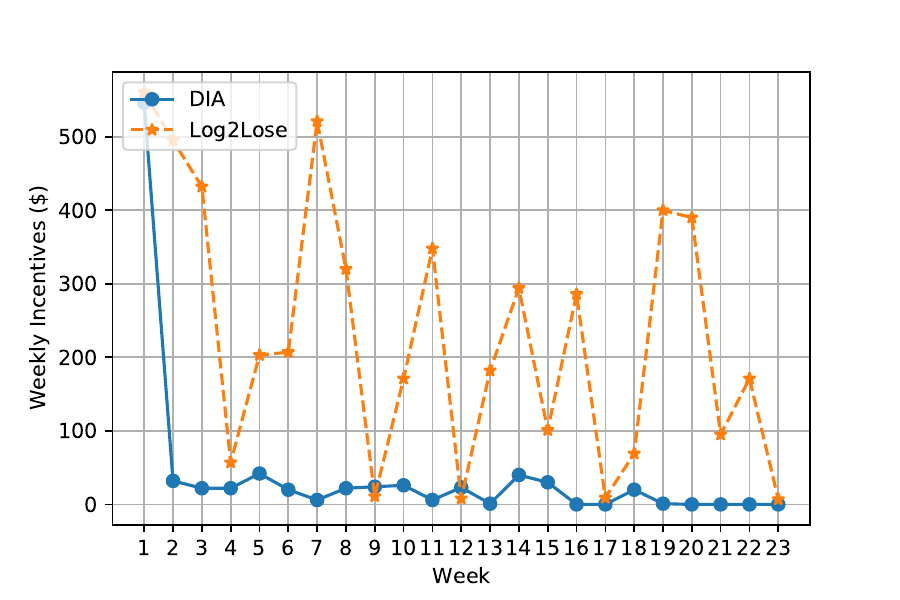}
          \caption{}
         \label{fig:weekly budget}
     \end{subfigure}
    
        \caption{Figure \ref{fig:cumulative budget} shows the cumulative incentives distributed using DIA with 100\% budget and the cumulative incentives distributed in the original Log2lose study. Figure \ref{fig:weekly budget} shows the incentives distributed per week using DIA and the incentives distributed in the original Log2lose study.}
        \label{fig:moderate reward}
\end{figure}

\subsection{Managerial insights}
%\ym{write something here about how healthcare providers could use our findings to make decisions}
%Consider our findings in the numerical experiments, we provide the following suggestions to help healthcare providers design efficient and effective financial incentives. 
Our work provides several key insights to both the operators of Log2Lose and healthcare providers who would implement financial incentive-based interventions for weight loss.
\begin{enumerate}
\item \emph{Spending more of the budget on incentives early in the intervention improves outcomes.}  The goal of providing financial incentives is to increase and maintain a participant's internal and external motivation for weight loss, and thus increase adherence to the intervention. Using our behavioral framework, we find that distributing larger incentives at the beginning helps increases a participant's external motivation, which increases a participant's weight loss for a moderate amount of time. Although such external motivation fades quickly, consistent weight loss success at the early stages can help increase the internal motivation, which has a long-lasting effect on increasing participant's adherence to the program. As a result, Figure \ref{fig:hinge cumulative} and \ref{fig:hinge marginal} show that participants need significantly less incentives in the later weeks, and the surplus can be distributed to those who need extra incentive to increase their adherence to the program.

Our findings are consistent with predictions from human behavioral literature, which suggests that larger incentives are more effective for individuals who are only starting to modify their behavior to induce behavioral changes and less effective for people who successfully incorporate the new behaviors into their lifestyle \citep{gneezy2011and,springer2016designing}. The Log2Lose study attempted to provide greater rewards by
providing participants with \$10 during each of the first four weeks, if they met incentive criteria (i.e., logged enough calories and/or lost weight, depending on randomization assignment). Our approach expands the Log2Lose approach by allowing greater amounts initially so as to increase extrinsic motivation, and thus weight loss. 

\item \emph{Moderate reward yielded large behavioral impact.} Once a significant amount of budget is distributed in the first weeks of the program, we find less incentives are required to keep participants losing weight. This eventually leads to moderate incentive reward on average across the entire intervention. The results in Figure \ref{fig:cumulative budget} and \ref{fig:weekly budget} show \$3-\$4 on average per week per participant is sufficient for helping the entire group of participants to achieve 5\% weight loss. Moderate rewards may be sufficient after initial weight loss as participants' motivation becomes more intrinsic as they succeed with weight loss. The greater intrinsic motivation may be sufficient to sustain their weight loss efforts throughout the trial,
%the duration of the trial.

%\ym{YM: again relate this more to the features of the model/algorithm and why they make this possible, ALSO: would be good to add some idea of \$ spent per participant here to give a sense of scale (don't need a new figure for this since its all in the plot) or some notion of \% weight loss/ \$ spent / participant }

\item \emph{Intermittent rewards provide longer term benefit (i.e. implement a stochastic incentive policy over a deterministic one).} 
%\ym{YM: why would this be true given our approach? also harp on how this matches insights from the behavioral psychology literature and why this is good our model can capture this}
 Our findings are consistent with research on reinforcement schedules and match the insights from the behavioral literature in that a random incentive scheme can induce more human efforts \citep{ederer2013gaming}. In our simulation study, random incentives lead to increasing numbers of participants achieving clinically significant weight loss success. The results in Figure \ref{fig:budget_patient} and Figure \ref{fig:budget_percent} show a randomized policy has the potential to outperform a deterministic policy. In particular, by implementing a randomized policy participants lose higher percentage of their initial weight. 
 %These imply that although financial incentives have immediate impacts on a participant's weight loss, our long-term goal is to increase a participant's internal motivation by distributing incentives to encourage necessary behavioral changes. 
 %Therefore, a participant's long-term weight loss journey does not rely on consistent financial incentives, instead it is a result of high internal motivation and moderate external motivation.  Since financial incentives contribute less to an individual's weight loss as time progresses, deterministic incentive policy may not be the ideal policy. 
\end{enumerate}

\section{Conclusion}
In this paper, we develop a behavioral framework to design efficient and effective personalized financial incentives to help a large number of participants achieving clinical weight loss success. This framework includes a behavioral model to describe the weekly decision process of a participant, a surrogate maximum likelihood estimation model for estimating model parameters, and an algorithm to optimize personalized financial incentives with limited budget. Under deterministic incentive policy, we show our estimated incentives converge to the optimal incentives which can be computed assuming we have full knowledge of each participant. Furthermore, we evaluate the performance of our personalized incentive design. The results show our approach outperforms existing machine learning methods in predicting weight loss success, and increases weight loss success with significantly less budget. In terms of healthcare practices, our framework can be applied to design personalized financial incentives, and it can be implemented with any deterministic or stochastic incentive policy for clinical weight loss programs. 

 % outcomment this line in Case 2

%If you don't use BiBTex, you can manually itemize references as shown below.
%\newpage

%bibtex
%\bibliographystyle{informs2014}
%\bibliography{arxive_submission}

%% Here starts the e-companion (EC)
%%%%%%%%%%%%%%%%%%%%%%%%%%%%%%%%%%%%%%%%%%%%%%%%%%%%%%%%%%
\ECSwitch

%\ECDisclaimer
%%%%%%%%%%%%%%%%%%%%%%%%%%%%%%%%%%%%%%%%%%%%%%%%%%%%%%%%%%

%%% Main head for the e-companion
\ECHead{Appendices}
\begin{APPENDICES}
\section{Complete MILP formulation of SMLE problem} \label{sec: complete milp of SMLE}
 \begin{align}
       \min \quad &\alpha\sum_{t \in \mathcal{T}, d \in \mathcal{D}_t}|w_{t,d}-\tilde{w}_{t,d}|+\beta\sum_{t \in \mathcal{T}}|p_t - g_t|. \label{eq: SMLE 0}\\
\text{s.t.} \quad 
& w_{t,d+1}=bw_{t,d}+cf_{t,d+1}+k,  & t\in \{0,\cdots,23\}, \quad d \in \{0,\cdots,6\}  \label{eq: SMLE 1}\\
&f_{t,d}=dc_{t,d}+\xi_d, & t\in \{0,\cdots,23\}, \quad d \in \{0,\cdots,6\} \\
&f_{b,t+1} = \gamma_f f_{b,t} + (1-\gamma_f)\sum_{d=0}^6 f_{t,d}, &  t\in \{0,\cdots,23\}\\
%& \tilde{w}_{w,d} = w_{w,d}  + (1-d_{w,d})\zeta_t, \quad d \in [0,1,\cdots,6]\\
&c_{t,d}= f_{b,t}-\frac{a_{1,t}c\sum_{i=6-d}^6 b^i}{2}-\frac{a_{2,t}\hat{r}^w_{t}cb^{6-d}}{4A} & t\in \{0,\cdots,23\}, \quad d \in \{0,\cdots,6\} \\
&p_{t+1}=\gamma_p(p_t-p_{b})+p_{b}+k_pg_t,  &t\in \{0,\cdots,23\}\\
&w_{t,0}- w_{t,6} \leq M_{1,t} (1-l_{1,t}), \quad  &t \in \{0, \cdots, 23\}\\
	&p_t-B \leq M_{2,t}l_{2,t},  &t \in \{0, \cdots, 23\}\\
	& z_{1,t} \leq M_{z1}l_{2,t},   &t \in \{0, \cdots, 23\}\\
	&z_{1,t} \leq r^c_t, &  t \in \{0, \cdots, 23\}\\
	&z_{1,t} \geq r^c_t - M_{z1}(1-l_{2,t}), & t \in \{0, \cdots, 23\}\\
	& z_{1,t} \geq 0, &  t \in \{0, \cdots, 23\}\\
	& z_{3,t} \leq M_{z3}l_{1,t}, &  t \in \{0, \cdots, 23\}\\
	&z_{3,t} \leq k_1, &  t \in \{0, \cdots, 23\}\\
	&z_{3,t} \geq k_1-M_{z3}(1-l_{1,t}), &  t \in \{0, \cdots, 23\}\\
	& z_{3,t} \geq 0, & t \in \{0, \cdots, 23\}\\
	&a_{1,t+1} = \gamma_1(a_{1,t} - a_{1,b}) + a_{1,b} + z_{1,t} + z_{3,t}, & t \in \{0,\cdots,23\}\\
& z_{2,t} \leq M_{z2}l_{1,t}, & t \in \{0, \cdots, 23\}\\
	&z_{2,t} \leq k_2, &  t \in \{0, \cdots, 23\}\\
	&z_{2,t} \geq k2 - M_{z2}(1-l_{1,t}), & t \in \{0, \cdots, 23\}\\
	& z_{2,t} \geq 0, &  t \in \{0, \cdots, 23\}\\
	& a_{2,t+1} = \gamma_2(a_{2,t} - a_{2,b}) + a_{2,b} + r^w_t z_{2,t}, & t\in \{0,\cdots,23\}\\
	& l_{1,t}, l_{2,t} \in \{0,1\} & t\in \{0,\cdots,23\}\label{eq: SMLE 2}
\end{align}

 \section{Proofs of propositions in text}  
 \subsection{Proof of Proposition \ref{prop: mle formulation}}\label{sec: mle formulation}
To formulate this MLE problem recall that $\tilde{w}_{t,d} = w_{t,d} + \epsilon_{t,d}$, where $\epsilon_{t,d}$ are i.i.d.  $\epsilon_{t,d} \sim \text{Laplace}(0,\sigma)$. Recalling from Section  \ref{sec:pat_bet_wek_mod} that $g_t \sim \text{Bernoulli}(p_t)$, and letting $\mathcal{T}$ be the index set of all weeks in the study and $\mathcal{D}_t$ be the set of days during week $t\in \mathcal{T}$ that have weight observations,  we can expand the joint likelihood function as follows:  
 %
 %In Section \ref{sec:pat_prob} we have developed the in-week and between-week models that describe the time-varying dynamics of different states. Since the values of these states cannot be directly observed, we first attempt to estimate the values of these states using maximum likelihood estimation (MLE) and then modify it as a surrogate maximum likelihood (SMLE) model, which can be solved using any optimization solvers. To do this, we use observations $\hat{w}_{t,d}$, which denotes the weight record on day $d$ on week $t$, and $g_{t}$ which is a binary variable indicating whether or not the calorie recording is satisfied. Let $T$ be the index set for week and $D$ be the index set for day, the likelihood function over the parameter space $\mathbb{P}(\{\hat{w}_{t,d}\}_{(t,d) \in T \times D},\{g_t\}_{t\in T}| \{w_{t,d}, c_{t,d}\}_{(t,d) \in T \times D }\{a_{1,t}, a_{2,t}, p_{t}, f_{b,t}\}_{t \in T}) =$
 \begin{equation}
     \begin{aligned}
       & \mathbb{P}(\{\tilde{w}_{t,d},g_t\}_{t \in \mathcal{T}, d \in \mathcal{D}_t}| \{w_{t,d}, c_{t,d},a_{1,t}, a_{2,t}, p_{t}, f_{b,t},\hat{r}^w_t,B,k_1,k_2,k_p,r^w_t,r^c_t\}_{(t,d) \in \mathcal{T} \times [0,...,6] }) = \\& \prod_{t\in \mathcal{T}}  \big(\mathbb{P}(g_t|p_t)\mathbb{P}(f_{b,t}|f_{b,t-1},\{c_{t,d}\}_{d \in D})
      \mathbb{P}(a_{1,t}| a_{1,t-1},a_{1,b},k_1,r_t^c, p_t,\{w_{t,d}\}_{d=0}^6,B)\\
      &\mathbb{P}(a_{2,t}|a_{2.t-1},a_{2,b},k_2,r_t^w,\{w_{t,d}\}_{d=0}^6)\mathbb{P}(p_t|p_{t-1},p_b,k_p,g_{t-1})\\
       &\mathbb{P}(\tilde{w}_{t,0}|w_{t,0})\mathbb{P}(w_{t,0}|w_{t-1,6},c_{t,0})
     \mathbb{P}(c_{t,0}|w_{t-1,6},a_{1,t},a_{2,t},f_{b,t},\hat{r}^w_{t}) \big)\\ 
      &\prod_{t \in \mathcal{T}, d \in \mathcal{D}_t} \mathbb{P}(\tilde{w}_{t,d}|w_{t,d})\prod_{(t,d) \in \mathcal{T} \times [0,...,6]} \big(\mathbb{P}(w_{t,d}|w_{t,d-1},c_{t,d})
     \mathbb{P}(c_{t,d}|w_{t,d-1},a_{1,t},a_{2,t},f_{b,t},\hat{r}^w_{t}) \big)
     \end{aligned}
     \label{eq:joint_liklihood_full}
 \end{equation}
Note that many of the terms in the joint likelihood function are in fact degenerate distributions by the assumptions of the model in Section \ref{sec:pat_prob}. Thus by taking the log of the above expression and expressing degenerate distributions as deterministic constraints we get the desired formulation.
\halmos
\endproof
 
 \subsection{Proof of Proposition \ref{prop:pat_prob_opt}}\label{sec: induction proof}
 
To prove the proposition we will solve the in week problem explicitly with dynamic programming. Let $V_{t,6}(w_{t,j})$ be the value function of a sub-problem maximizing the utility function from day $j \in \{0,..,5\}$ to the end of day 6 (Sunday) of week $t$. We want to show that:
 \begin{equation}
 \begin{aligned}
    V_{t,6}(w_{t,j})&=\max_{c_{t,j}} -a_{1,t}\left((\sum_{i=0}^{6-j}b^{i+1})w_{t,j-1}+(\sum_{d=j}^6(\sum_{i=0}^{6-d}cb^i)c_{t,j})+(\sum_{d=j}^6(\sum_{i=0}^{6-d}b^i)k)\right)\\
    &+a_{2,t}\hat{r}^w_{t}\frac{w_{t,\bar{d}}-b^{5-j}w_{t,j-1}-\sum_{d=j}^6 (cb^{6-d}c_{t,d})-\sum_{d=j}^6(b^{6-d}k)+A}{2A}\\
    &-\sum_{d=j}^6 (c_{t,d}^2-2c_{t,d}f_{b,t}+\mathbb{E}[\xi_{t,d}^2]+f_{b,t}^2)
    \end{aligned}
    \label{eq:pat_prob_val_struct}
 \end{equation}
 Because, if \eqref{eq:pat_prob_val_struct} is the correct structure, then the sub problems of the in-week model can be written as a sequence of convex optimization problems.
 First consider the base case $j=5$:
\begin{equation}
\begin{aligned}
 V_{t,6}(w_{t,5}) = &\max_{c_{t,6}} \mathbb{E}[-a_{1,t}(bw_{t,5}+c(c_{t,6}+\xi_{t,5})+k)+a_{2,t}\hat{r}^w_{t}\mathbbm{1}\{w_{t,0}-w_{t,6}>0\}-(c_{t,6}+\xi_{t,6}-(f_{b,t})^2]\\
    =&\max_{c_{t,6}} -a_{1,t}(bw_{t,5}+cc_{t,6}+ck)+a_{2,t}\hat{r}^w_{t} \mathbb{P}(w_{t,0}-w_{t,6}>0)- (c_{t,6}^2-2c_{t,6}f_{b,t}+\mathbb{E}[\xi_{t,6}^2]+f_{b,t}^2)
\end{aligned}
\label{eq:in_week_bc}
\end{equation}

 Since $\mathbb{P}(w_{t,0}-w_{t,6}>0)=\mathbb{P}(\xi_{t,6} \leq \frac{w_{t,0} - bw_{t,5} -cc_{t,6} -k}{c})$ and
 $\xi_{t,6} \sim U(-A,A)$, $\mathbb{P}(\xi_{t,6} \leq \frac{w_{t, 0} - bw_{t,5} -cc_{t,6} -k}{c})=\frac{w_{t, 0}-bw_{t,5}-cc_6-k+A}{2A}$. Substituting this into \eqref{eq:in_week_bc}:
 \begin{equation}
     \begin{aligned}
     V_{t,6}(w_{t,5})= \max_{c_{t,6}}& -a_{1,t}(bw_{t,5}+cc_{t,6}+ck)+a_{2,t}\hat{r}^w_{t}\frac{w_{t, 0}-bw_{t,5}-cc_6-k+A}{2A}\\
     &-(c_{t,6}^2-2c_{t,6}f_{b,t}+\mathbb{E}[\xi_{t,6}^2]+f_{b,t}^2)
     \end{aligned}
     \label{eq:val_func_base}
 \end{equation}
 This proves the base case since \eqref{eq:val_func_base} follows the desired form. Note, that this is a concave quadratic optimization problem, so a stationary point will be a global optimal solution. Next we take the derivative of the equation with respect to $c_{t,6}$ and set it equal to $0$, we find the optimal solution $c_{t,6}^*=f_{b,t}-\frac{a_{1,t}c}{2}-\frac{a_{2,t}\hat{r}^w_{t}c}{4A}$.
 
 %Next, find the optimal solution of $c_{t,5}^*$ for $V_{t,6}(w_{t,4})=\max_{c_5} -a_{1,t}(bw_{t,4}+cc_{t,5})-a_4(c_{t,5}^2+\xi_{t,5}^2+f_{b,t}^2)+ V_6(w_{t,5})$. 
 
 Next we make the following inductive hypothesis for some $0\leq j<6$:
 
 \begin{equation}
     \begin{aligned}
     V_{t,6}(w_{t,j})&=\max_{c_{t,j+1}} -a_{1,t}\left((\sum_{i=0}^{6-j-1}b^{i+1})w_{t,j}+(\sum_{d=j+1}^6(\sum_{i=0}^{6-d}cb^i)c_{t,j+1})+(\sum_{d=j+1}^6(\sum_{i=0}^{6-d}b^i)k)\right)\\
    &+a_{2,t}\hat{r}^w_{t}\frac{w_{t,0}-b^{5-j-1}w_{t,j}-\sum_{d=j+1}^6 (cb^{6-d}c_{t,d})-\sum_{d=j+1}^6(b^{6-d}k)+A}{2A}\\
    &-\sum_{d=j+1}^6 (c_{t,d}^2-2c_{t,d}f_{b,t}+\mathbb{E}[\xi_{t,d}^2]+f_{b,t}^2),
     \end{aligned}
 \end{equation}
 
 Then the $V_{t,6}(w_{t,j-1})$ can be computed as:
 \begin{equation}
     \begin{aligned}
      V_{t,6}(w_{t,j-1}) = \max_{c_{t,j}} &-a_{1,t}w_{t,j}-(c_{t,j}+\xi_j-f_{b,t})^2+ V_{w,6}(w_{t,j})\\
      =\max_{c_{t,j}}  &-a_{1,t}(bw_{t,j-1}+cc_{t,j}+k)-(c_{t,j}^2-2c_{t,j}f_{b,t}+\xi_{j}^2+f_{b,t}^2)\\
      & -a_{1,t}\left((\sum_{i=0}^{6-j-1}b^{i+1})w_{t,j}+(\sum_{d=j+1}^6(\sum_{i=0}^{6-d}cb^i)c_{t,j+1})+(\sum_{d=j+1}^6(\sum_{i=0}^{6-d}b^i)k)\right)\\
      &+a_{2,t}\hat{r}^w_{t}\frac{w_{t,0}-b^{5-j-1}(bw_{t,j-1}+cc_{t,j}+k)-\sum_{d=j+1}^6 (cb^{6-d}c_{t,d})-\sum_{d=j+1}^6(b^{6-d}k)+A}{2A}\\
      & -\sum_{d=j+1}^6 (c_{t,d}^2-2c_{t,d}f_{b,t}+\xi_{t,d}^2+f_{b,t}^2)\\
       =\max_{c_{w,j}} &-a_{1,t}\left((\sum_{i=0}^{6-j}b^{i+1})w_{t,j-1}+(\sum_{d=j}^6(\sum_{i=0}^{6-d}cb^i)c_{t,j})+(\sum_{d=j}^6(\sum_{i=0}^{6-d}b^i)k)\right)\\
    &+a_{2,t}\hat{r}^w_{t}\frac{w_{t,0}-b^{5-j}w_{t,j-1}-\sum_{d=j}^6 (cb^{6-d}c_{t,d})-\sum_{d=j}^6(b^{6-d}k)+A}{2A}\\
    &-\sum_{d=j}^6 (c_{t,d}^2-2c_{t,d}f_{b,t}+\xi_{t,d}^2+f_{b,t}^2)
     \end{aligned}
     \label{eq:in_week_final_induct}
 \end{equation}
Which proves our claim that the structure of \eqref{eq:pat_prob_val_struct} holds for all days of the week as desired. To complete the proof and show that $c_{i,j}^*$ has the desired form, we can take the derivative of \eqref{eq:in_week_final_induct} with respect to $c_{t,j}$ and set it equal to 0, which yields $c_{t,j}^* = f_{b,t}-\frac{a_{1,t}c\sum_{i=6-j}^6 b^i}{2}-\frac{a_{2,t}\hat{r}^w_t cb^{6-j}}{4A}$ as desired.
\halmos

\endproof

\subsection{Proof of Proposition \ref{prop:reform_a_dynam}} \label{app:a1}
First we define two sets of binary variables $\{l_{1,t}\}_{t=0}^{23}$ and $\{l_{2,t}\}_{t=0}^{23}$. Using the Big-M technique \citep{wolsey1999integer}, let $l_{1,t}=1$ if $w_{t,6} < w_{t,0}$ and $l_{1,t}=0$ if $w_{t,6} \geq w_{t,0}$. Similarly, let $l_{2,t}=1$ if $p_t \geq B$ and $l_{2,t}=0$ if $p_t < B$. Constraint \ref{eq: a1 reform 1} enforces $l_{1,t}=1$ if $w_{t,0} < w_{t,6}$ and $l_{1,t}=0$ if $w_{t,0} \geq w_{t,6}$. Similarly, Constraint \ref{eq: a1 reform 2} enforces $l_{2,t}=1$ if $p_t \geq B$ and $l_{2,t}=0$ if $p_t  < B$. Constraint \ref{eq: a1 reform 3}-\ref{eq: a1 reform 5} is the reformulation of $r^c_t\mathbbm{1}\{p_w-B\geq 0\}$. If $l_{2,t}=0$, then $z_{1,t}=0$. If $l_{2,t}=1$, $z_{1,t}=r^c_t$ since Constraint \ref{eq: a1 reform 4} is a tighter upper bound for $z_{1,t}$ than Constraint \ref{eq: a1 reform 3}, and Constraint \ref{eq: a1 reform 5} ensures $z_{1,t}$ must be greater than or equal to $r^c_t$. Similarly, Constraint \ref{eq: a1 reform 7}-\ref{eq: a1 reform 10} ensures $z_{3,t}=0$ if $l_{1,t}=0$ and $z_{3,t}=k_1$ if $l_{1,t}=1$. Then in Constraint \ref{eq: a1 reform 11} we replace the nonlinear terms with $z_{1,t}$ and $z_{3,t}$. \halmos\endproof

 \subsection{Proof of Proposition \ref{prop:reform_a_2_dynam}} \label{app:a2}
 %\proof{Proof of Proposition \ref{prop:reform_a_2_dynam}}
Similar to the proof of Proposition \ref{prop:reform_a_dynam},  we use the Big-M technique \citep{wolsey1999integer} to reformulate nonlinear constraints as linear ones.
First we introduce the binary variables $l_{1,t}$, $l_{2,t}$, $z_{1,t}$, and $z_{2,t}$. Constraint \eqref{eq: a2 reform 1}  enforces $l_{1,t}=1$ if $w_{t,6} < w_{t,0}$ and $l_{1,t}=0$ if $w_{t,6} \geq w_{t,0}$. Constraint \eqref{eq: a2 reform 2}-\eqref{eq: a2 reform 5} are the reformulation of $k_2 \mathbbm{1}\{(w_{t,0}- w_{t,6})>0\}$, which indicates $z_{2,t}=0$ if $l_{1,t}=0$ and $z_{2,t}=k_2$ if $l_{1,t}=1$. Constraint \ref{eq: a2 reform 2}-\ref{eq: a2 reform 3} ensures $z_{2,t} \leq k_2$, and Constraint \ref{eq: a2 reform 4} ensures $z_{1,t}$ must be greater than or equal to $r^c_t$. Lastly, in Constraint \ref{eq: a2 reform 6} the product of the binary and the continuous variables is replaced with $z_{2,t}$.  \halmos \endproof

\subsection{Proof of Proposition \ref{prop:surogate}} \label{app:surogate}
We prove the inequalities hold for $g_t=0$ and $g_t=1$ separately. 
%First, notice the likelihood function $P(g_t|p_t)=(p_t)^{g_t}(1-p_t)^{1-g_t}$  and the log-likelihood function $-\log(P(g_t|p_t))=-\log(p_t^{g_t}(1-p_t)^{1-g_t})=-g_t\log(p_t)-(1-g_t)\log(1-p_t)$.  

If $g_t = 0$, then the log-likelihood function $-\log(\mathbb{P}(g_t=0|p_t))=-\log(p_t^{g_t}(1-p_t)^{1-g_t})=-g_t\log(p_t)-(1-g_t)\log(1-p_t)=-\log(1-p_t)$ and $|g_t-p_t|=|0-p_t|=p_t$. Let $f:[\epsilon,1-\epsilon] \mapsto \mathbb{R}$ be: $f(x) = \frac{-\log(1-x)}{x}p_t + \log(1-p_t)$. Note, $f(p_t) = 0$. Computing the first derivative of $f$ yields $\frac{df}{dx} =\frac{\frac{x}{1-x}+\log(1-x)}{x^2}$. Note that $\frac{df}{dx} > 0$, meaning $f$ is monotonically increasing in $x$, and thus $f(\epsilon) \leq f(p_t) \leq f(1-\epsilon)$ which gives the desired inequalities.
%
%
%Let $\epsilon \leq x \leq 1-\epsilon$, then $-\log(1-p_t)=\frac{-\log(1-x)}{x}p_t$ when $x=p_t$. Next, we compute the first derivative of $\frac{-\log(1-x)}{x}$ and determine whether the function is monotonically increasing or monotonically decreasing. By taking the first derivative, we get $ \frac{d}{dx} \frac{-\log(1-x)}{x}=\frac{\frac{x}{1-x}+\log(1-x)}{x^2}$. Since the first derivative is strictly positive for $\epsilon \leq x \leq 1-\epsilon$, the function $\frac{-\log(1-x)}{x}$ is monotonically increasing. Therefore, the inequalities $-\log(\mathbb{P}(g_t=0|p_t)) \leq \frac{-log(\epsilon)}{1-\epsilon}p_t$ and $-\log(\mathbb{P}(g_t=0|p_t)) \geq \frac{-\log(1-\epsilon)}{\epsilon} p_t$ hold true for $g_t=0$.

If $g_t = 1$, then $\log(\mathbb{P}(g_t=1|p_t))=-\log(p_t)$ and $|g_t-p_t|=1-p_t$. Define $h:[\epsilon,1-\epsilon] \mapsto \mathbb{R}$ as $h(x) = \frac{-\log(x)}{1-x}(1-p_t) + \log(p_t)$, note $h(p_t) = 0$. We can compute the first derivative of $h$ as $ \frac{dh}{dx} =\frac{x+x(-\log(x))-1}{x(1-x)^2}$, and note that $\frac{dh}{dx} < 0$ meaning $h$ is monotonically decreasing. Therefore,   $h(1-\epsilon) \leq h(p_t) \leq h(\epsilon)$ which provides the desired result.\halmos \endproof

% Let $\epsilon \leq x \leq 1-\epsilon$, then 
% $-\log(p_t)=\frac{-\log(x)}{1-x}(1-p_t)$ when $x=p_t$. To determine whether the function $\frac{-\log(x)}{1-x}$ is monotonically increasing or decreasing, we take its first derivative: $ \frac{d}{dx} \frac{-\log(x)}{1-x}=\frac{x+x(-\log(x))-1}{x(1-x)^2}$. Notice the first derivative is strictly negative for $\epsilon \leq x \leq 1-\epsilon$, which implies the function $\frac{-\log(x)}{1-x}$ is monotonically decreasing. Therefore, the inequalities $-\log(\mathbb{P}(g_t=1|p_t)) \leq \frac{-log(\epsilon)}{1-\epsilon}(1-p_t)$ and $-\log(\mathbb{P}(g_t=1|p_t)) \geq \frac{-\log(1-\epsilon)}{\epsilon} (1-p_t)$ also hold true for $g_t=1$.

\subsection{Proof of Proposition \ref{prop:consistency}} \label{app:consistency}
 Let $(w_{0,0}^*,\theta_{0}^*)$ be the true initial conditions. Then for any possible initial conditions $(w_{0,0},\theta_{0}) \neq (w_{0,0}^*,\theta_{0}^*) $ we can express the surrogate posterior as follows: 
 \begin{equation}
 \begin{aligned}
     \log &(\hat{\mathbb{P}}(w_{0,0},\theta_{0}|\{\tilde{w}_{t,d},g_{t}, r^w_t, r^c_t\}))=\log(\hat{\mathbb{P}}(w_{0,0}^*,\theta_{0}^*|\{\tilde{w}_{t,d},g_{t}, r^w_t, r^c_t\}))\\
     &+ \sum_{t\in\mathcal{T},d \in \mathcal{D}_t} \log\frac{\mathbb{P}(\bar{w}_{t,d}-\tilde{w}_{t,d})}{\mathbb{P}(w_{t,d}-\tilde{w}_{t,d})} + \sum_{t \in \mathcal{T}}(|g_t-\bar{p}_t|-|g_t-p_t^*|)-\log \frac{\mathbb{P}(w_{0,0}, \theta_{0})}{\mathbb{P}(w_{0,0}^*,\theta_{0}^*)}
 \end{aligned} 
 \label{eq:surg_post_extend}
 \end{equation} 
 Using the results of Proposition \ref{prop:surogate}, we can bound $\sum_{t \in \mathcal{T}} (|g_t-p_t^*|-|g_t-\bar{p}_t|) \leq \epsilon_{\max}\sum_{t \in \mathcal{T}} \frac{\log(\mathbb{P}(g_t|p^*_t))}{\log(\mathbb{P}(g_t|\bar{p}_t))}$ , where 
 $\epsilon_{\max} = \max \{\frac{\epsilon}{\log(1-\epsilon)}, \frac{1-\epsilon}{\log(\epsilon)}\}$. Thus we see: 

 \begin{multline}
     \eqref{eq:surg_post_extend}  \leq \log(\hat{\mathbb{P}}(w_{0,0}^*,\theta_{0}^*|\{\tilde{w}_{t,d},g_{t}, r^w_t, r^c_t\}))\\
     + \sum_{t\in\mathcal{T},d \in \mathcal{D}_t} \log\frac{\mathbb{P}(\bar{w}_{t,d}-\tilde{w}_{t,d})}{\mathbb{P}(w_{t,d}-\tilde{w}_{t,d})} +  \epsilon_{\max}\sum_{t \in \mathcal{T}} \frac{\log(\mathbb{P}(g_t|p^*_t))}{\log(\mathbb{P}(g_t|\bar{p}_t))} -\log \frac{\mathbb{P}(w_{0,0}, \theta_{0})}{\mathbb{P}(w_{0,0}^*,\theta_{0}^*)}
  \end{multline}

 Since $\frac{\mathbb{P}(w_{0,0},\theta_{0})}{\mathbb{P}(w_{0,0}^*,\theta_{0}^*)}$ is a constant and $\log(\hat{\mathbb{P}}(w_{0,0}^*,\theta_{0}^*|\{\tilde{w}_{t,d},g_{t}, r^w_t, r^c_t\})) \in [0,1]$ by definition,
then combined with Assumption \ref{as:sufficient_excite}  this implies $\max_{S(\delta)} \log (\hat{\mathbb{P}}(w_{0,0},\theta_{0}|\{\tilde{w}_{t,d},g_{t}, r^w_t, r^c_t\})) \rightarrow -\infty \text{ } \forall \delta>0$. This implies $\max_{S(\delta)} \hat{\mathbb{P}}(w_{0,0},\theta_{0}|\{\tilde{w}_{t,d},g_{t}, r^w_t, r^c_t\}) \rightarrow 0$. 

To complete the proof consider the probability mass placed on $S(\delta)$ given by $\hat{\mathbb{P}}(S(\delta)|\{\tilde{w}_{t,d}, g_t, r^w_t, r^c_t\})=\int_{S(\delta)} \hat{\mathbb{P}}(w_{0,0},\theta_{0}|\{\tilde{w}_{t,d},g_{t}, r^w_t, r^c_t\})  dw_{0,0}d\theta_0 \leq \text{Vol}(\mathcal{W}\times \mathcal{\theta}) \max_{S(\delta)} \hat{\mathbb{P}}(w_{0,0},\theta_{0}|\{\tilde{w}_{t,d},g_{t}, r^w_t, r^c_t\}) \rightarrow 0$. Thus our surrogate posterior meets the definition as desired.\halmos 
\endproof
\subsection{Proof of Corollary \ref{cor:consist}}
 Since the event $\big\{(\hat{w}^{MAP}_{0,0},\hat{\theta}^{MAP}_0) \notin \mathcal{B}((w^*_{0,0}, \theta^*_{0}), \delta)) \big\}$ is a subset of the event $\big\{\max_{S(\delta)} \hat{\mathbb{P}}(w_{0,0},\theta_{0}|\{\tilde{w}_{t,d},g_{t}, r^w_t, r^c_t
\}) \geq \max_{w_{0,0},\theta_0 \in \mathcal{B}((w^*_{0,0}, \theta^*_{0}), \delta)} \hat{\mathbb{P}}(w_{0,0},\theta_{0}|\{\tilde{w}_{t,d},g_{t}, r^w_t, r^c_t
\}) \big\}$, which implies $\mathbb{P}((\hat{w}^\text{MAP}_{0,0},\hat{\theta}^\text{MAP}_0) \notin \mathcal{B}((w^*_{0,0},\theta^*_{0}), \delta)) \leq \mathbb{P}(\max_{S(\delta)} \hat{\mathbb{P}}(w_{0,0},\theta_{0}|\{\tilde{w}_{t,d},g_{t}, r^w_t, r^c_t
\}) \geq \max_{w_{0,0},\theta_0 \in \mathcal{B}((w^*_{0,0}, \theta^*_{0}), \delta)} \hat{\mathbb{P}}(w_{0,0},\theta_{0}|\{\tilde{w}_{t,d},g_{t}, r^w_t, r^c_t
\}))$. By Proposition \ref{prop:consistency}, $\mathbb{P}(\max_{S(\delta)} \hat{\mathbb{P}}(w_{0,0},\theta_{0}|\{\tilde{w}_{t,d},g_{t}, r^w_t, r^c_t\})) \rightarrow 0 $ as $\mathcal{T} \rightarrow \infty$ and hence the result of the corollary follows. \halmos \endproof

\subsection{Proof of Proposition \ref{prop: semicont}} \label{app:semicont}
Propositions \ref{prop:reform_a_dynam} and \ref{prop:reform_a_2_dynam} indicate the problem described in \eqref{eq:incentive_feasibility} can be reformulated with a set of linear constraints which are affine in $(w_{u,0,0}, \theta_{u,0}\{r^w_{u,t}, r^c_{u,t}\}_{t=0}^T) \forall u \in \mathcal{U}$. This implies the function $ \psi(\{w_{u,0,0},\theta_{u,0},\{r^w_{u,t}, r^c_{u,t}\}_{t=0}^{T+n}\}_{u \in U})$ is lower semi-continuous to each argument by applying results from \citep{hassanzadeh2014generalization}.
\halmos  \endproof  

\subsection{Proof of Proposition \ref{prop: semicont approx}} \label{app:semicont approx}
Corollary \ref{cor:consist} implies the surrogate posterior estimates $\hat{\mathbb{P}}(w_{u,0,0},\theta_{u,0}|\{\tilde{w}_{u,t,d},g_{u,t},r^w_{t},r^c_{t}\}_{t=0}^T)$ are statistically consistent and Proposition \ref{prop: semicont} implies 
$ \psi(\{w_{u,0,0},\theta_{u,0},\{r^w_{u,t}, r^c_{u,t}\}_{t=0}^{T+n}\}_{u \in U})$ is lower semi-continuous in all of its arguments. Hence by Proposition 2.1.ii of \cite{lachout2005strong}  $ \psi (\{\hat{w}^T_{u,0,0},\hat{\theta}^T_{u,0},\{\bar{r}^w_{u,t}, \bar{r}^c_{u,t}\}_{t=0}^{24}\}_{u\in U}) $ is a lower semi-continuous approximation of the function $\psi (\{w^*_{u,0,0},\theta^*_{u,0},\{\bar{r}^w_{u,t}, \bar{r}^c_{u,t}\}_{t=0}^{24}\}_{u\in U})$ with respect to the true initial conditions.
\halmos\endproof

\subsection{Proof of Theorem \ref{thm:asymp opt}} \label{app:asymp opt}
Since Corollary \ref{cor:consist} implies $(\hat{w}^\text{MAP}_{0,0},\hat{\theta}_0^\text{MAP})\overset{p}{\to}(w^*_{0,0},\theta_0^*)$  and Proposition \ref{prop: semicont approx} implies $ \psi (\hat{w}^T_{u,0,0},\hat{\theta}^T_{u,0},\{\bar{r}^w_{u,t}, \bar{r}^c_{u,t}\}_{t=0}^{24}\}_{u\in U})$ is a lower semi-continuous approximation to the function $\psi (\{w^*_{u,0,0},\theta^*_{u,0},\{\bar{r}^w_{u,t}, \bar{r}^c_{u,t}\}_{t=0}^{24}\}_{u\in U})$, the result follows Theorem 4.3 of (\cite{lachout2005strong}) which implies any solution $\{r^{w,DIA}_{u,T}, r^{c,DIA}_{u,T}\}_{u \in U}$ returned by Algorithm \ref{alg:incentive} are asymptotically optimal.
\halmos\endproof

 \end{APPENDICES}

%%%%%%%%%%%%%%%%%
\end{document}